\documentclass[times,twocolumn,final]{elsarticle}
\usepackage{medima}
\usepackage{framed,multirow}

\usepackage{amssymb}
\usepackage{latexsym}

\usepackage{url}
\usepackage{xcolor}
\usepackage{amsmath,amssymb,amsfonts}
\usepackage{algorithm,algorithmicx}
\usepackage[noend]{algpseudocode}
\usepackage{graphicx}
\usepackage{textcomp}
\usepackage{multirow}
\usepackage{bm}
\usepackage{url}
\usepackage{hyperref}
\usepackage{color}
\usepackage{rotating}
\usepackage{booktabs}
\usepackage{xr}
\usepackage{caption}
\usepackage{textcomp}
\usepackage{gensymb}
\usepackage{float}

\usepackage{hyperref}
\newcommand{\modelname}{KeyMorph}

\newcommand{\moveimg}{\bm{x}_m}
\newcommand{\fiximg}{\bm{x}_f}
\newcommand{\alignimg}{\bm{x}_r}

\newcommand{\opttran}{\theta^*}
\newcommand{\norm}[1]{\left\lVert#1\right\rVert}
\newcommand{\nnparam}{w}
\newcommand{\nn}{f_\nnparam}

\DeclareMathOperator*{\argmin}{arg\,min}
\newcommand{\bpara}[1]{\vspace{0.3cm} \noindent \textbf{#1}}

\definecolor{newcolor}{rgb}{.8,.349,.1}

\journal{Medical Image Analysis}

\begin{document}

\verso{Alan Wang \textit{et~al.}}

\begin{frontmatter}

\title{A robust and interpretable deep learning framework for multi-modal registration via keypoints}%
% \tnotetext[tnote1]{This is an example for title footnote coding.}

\author[1,2]{Alan Q. Wang\corref{cor1}}
\ead{aw847@cornell.edu}
\cortext[cor1]{Corresponding author:}
\author[3]{Evan M. Yu}
\author[4,5]{Adrian V. Dalca}
\author[1,2]{Mert R. Sabuncu}

\address[1]{School of Electrical and Computer Engineering, Cornell University and Cornell Tech, New York, NY 10044, USA}
\address[2]{Department of Radiology, Weill Cornell Medical School, New York, NY 10065, USA}
\address[3]{Iterative Scopes, Cambridge, MA 02139, USA}
\address[4]{Computer Science and Artificial Intelligence Lab at the Massachusetts Institute of Technology, Cambridge, MA 02139, USA}
\address[5]{A.A. Martinos Center for Biomedical Imaging at the Massachusetts General Hospital, Charlestown, MA 02129, USA}

\received{April 05, 2023}
% \finalform{NA}
\accepted{August 23, 2023}
% \availableonline{NA}
% \communicated{S. Sarkar}

\begin{abstract}
%%%
We present \modelname, a deep learning-based image registration framework that relies on automatically detecting corresponding keypoints. 
State-of-the-art deep learning methods for registration often are not robust to large misalignments, are not interpretable, and do not incorporate the symmetries of the problem.
In addition, most models produce only a single prediction at test-time. 
Our core insight which addresses these shortcomings is that corresponding keypoints between images can be used to obtain the optimal transformation via a differentiable closed-form expression. 
We use this observation to drive the end-to-end learning of keypoints tailored for the registration task, and without knowledge of ground-truth keypoints. 
This framework not only leads to substantially more robust registration but also yields better interpretability, since the keypoints reveal which parts of the image are driving the final alignment. 
Moreover, \modelname~can be designed to be equivariant under image translations and/or symmetric with respect to the input image ordering.
Finally, we show how multiple deformation fields can be computed efficiently and in closed-form at test time corresponding to different transformation variants.
We demonstrate the proposed framework in solving 3D affine and spline-based registration of multi-modal brain MRI scans. 
In particular, we show registration accuracy that surpasses current state-of-the-art methods, especially in the context of large displacements. 
Our code is available at \texttt{https://github.com/alanqrwang/keymorph}.
%%%%
\end{abstract}


\end{frontmatter}

%\linenumbers

\begin{figure}[t]
\centering
\includegraphics[width=0.45\textwidth]{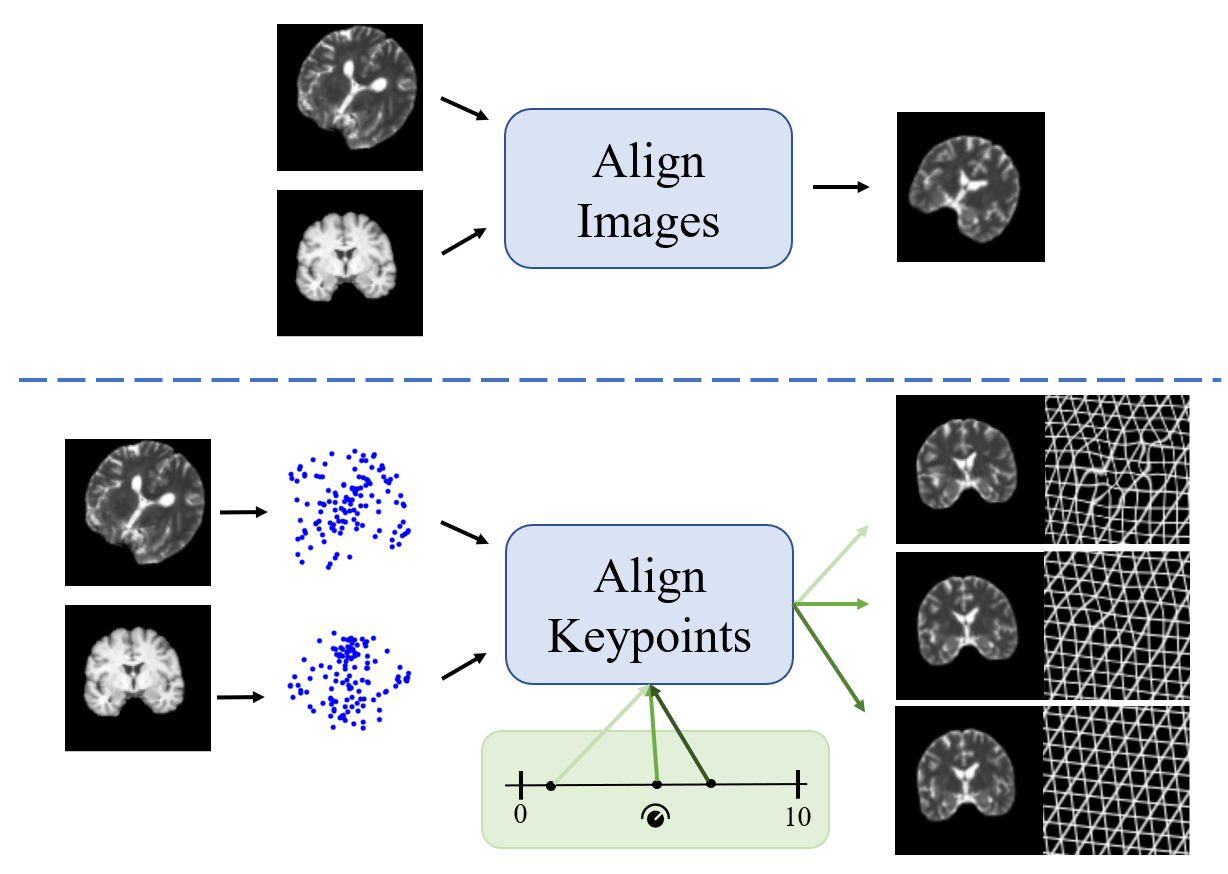}
   \caption{(Top) Existing registrations tools take as input a single moving and fixed image pair and output a single prediction by aligning images. These ``black-box" models are often not robust to large initial misalignments and are hard to interpret. (Bottom) \modelname~detects corresponding keypoints from input pairs and aligns the keypoints in an interpretable fashion. A knob allows for controllable user selection of many optimal registrations at test time corresponding to different degrees of nonlinearity.}
\label{fig:motivating}
\end{figure}
\section{Introduction}
\label{sec:introduction}
Registration is a fundamental problem in biomedical imaging tasks. Multiple images, often reflecting a variety of contrasts, are commonly acquired in many applications~\citep{uludaug2014general}. %, for example at different time points to track anatomical changes or study disease progression~\cite{uludaug2014general}. Image registration provides correspondences between scans, for example to analyze population variability or changes over time.
% Providing correspondences between separate medical imaging scans, known as image registration, is an essential part of clinical workflows and scientific studies. %, for example to analyze population variability or spatial changes over time. %This enables clinical evaluation and treatment, as well as the study of associations with attributes such as age or disease~\cite{yu2020learning, smith2002accurate}. 
% Most existing image analysis pipelines involve a core registration step between different modalities and/or across time~\cite{aljabar2009multi, yu2020auto}.
Classical (i.e. non-learning-based) registration methods involve an iterative optimization of a similarity metric over a space of transformations~\citep{oliveira2014medical,sotiras2013survey}. 
An optional regularization term scaled by a hyperparameter encourages smoothness of the transformation. 
Deep learning-based strategies leverage large datasets of images to solve registration and are able to perform fast inference via efficient feed-forward passes. 
These strategies use convolutional neural network (CNN) architectures that either output transformation parameters (e.g. affine or spline)~\citep{lee2019imageandspatial,de2019deep} or a dense deformation field~\citep{balakrishnan2019voxelmorph} which aligns an image pair. 

\begin{figure*}[t!]
\centering
\includegraphics[width=\textwidth]{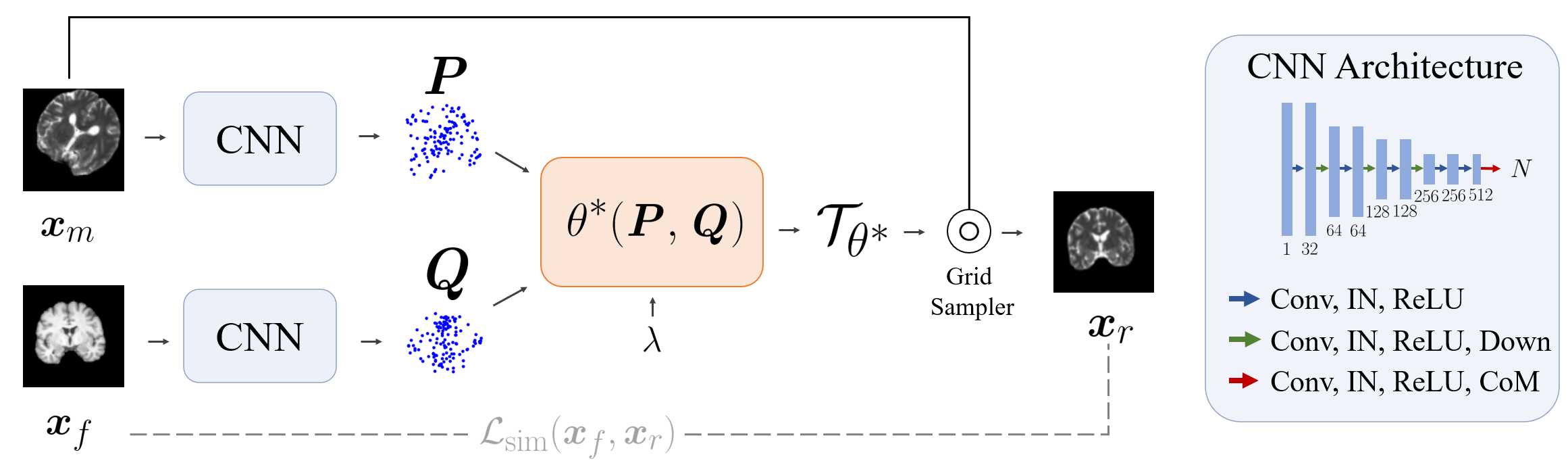}
   \caption{Proposed framework. Fixed and moving 3D images are passed through the same  keypoint detection network composed of convolutional layers, instance normalization (IN), downsampling (Down), and a final center-of-mass (CoM) layer that predicts $N$ keypoints useful for registration. The transformation parameters are then computed as a function of the keypoints, which are in turn used to resample the moving image.}
\label{fig:proposed_model}
\end{figure*}

However, we pinpoint several shortcomings of prior CNN-based methods:
\begin{itemize}
    \item They often fail when the given image pair has a large misalignment (see Fig.~\ref{fig:motivating}). 
    This is likely due to CNNs being unable to effectively learn long-range dependencies and correspondences. Existing systems typically require that image pairs are roughly aligned~\citep{balakrishnan2019voxelmorph,dalca2019unsupervised,de2019deep,qin2019}, or at least are in the same orientation~\citep{mocanu2021flowreg}.
    \item They lack interpretability, as they are essentially ``black-box models'' that output either transformation parameters or a deformation field and provide little insight into what drives the alignment.
    \item They generally do not exploit the symmetries and equivariances present in the problem. For example, if one of the images is translated by a fixed amount, this has a pre-determined effect on the optimal registration solution, and this property is not built into any of the architectures used today for image registration. Explicitly leveraging this inductive bias can improve performance and robustness.
\end{itemize}

% Separately, recent deep learning strategies have focused on producing a set of predictions (rather than a single prediction) at test time corresponding to different hyperparameter values which weight the smoothness prior~\cite{hoopes2021hypermorph,mok2021conditional}.
% These \textit{hyperparameter-agnostic} models condition on the hyperparameter by modulating the underlying architecture.
% This not only enables rapid hyperparameter tuning, but also allows the end user to leverage and exercise control over the rich and distinct information within and across each unique prediction.
% Thus, these controllable models put humans in the loop, an arguably important feature in learning models for medical imaging~\cite{Holzinger2016,xin2018accelerating}.
% However, in addition to the shortcomings previously discussed, a downside of these methods is that they require modulating the network in order to generate a prediction, which is computationally expensive.

\bpara{Contribution.} In this work, we present \modelname, an end-to-end deep learning-based framework that seeks to address these aforementioned issues. 
The main insight is that \textit{corresponding keypoints} can be used to derive the optimal transformation in closed-form, where the keypoints themselves are learned by a neural network.
These keypoints are optimized specifically for the registration task, and we do not assume knowledge of ground-truth keypoints.
Choosing a transformation whose optimal parameters can be solved in a differentiable manner given the learned keypoints enables end-to-end training of the registration pipeline.
In a sentence, rather than treating matched keypoint detection as a supervised learning problem requiring human-annotated keypoints, we propose to use an end-to-end unsupervised strategy tailored toward registration. 

This keypoint-based formulation is robust against misalignments as the closed-form solution is not sensitive to the initial position of the keypoints. 
Additionally, the model is interpretable since the keypoints that drive the alignment can be visualized. 
Furthermore, we also show how to incorporate symmetries into the model design. 
Finally, for a given transformation which accepts user-specified hyperparameters, a dense set of registrations can be computed at test-time directly from the learned keypoints, requiring no modulation of the neural network architecture.
For example, the architecture we describe in this paper is translation-equivariant and leverages a recently-proposed center-of-mass layer~\citep{ma2020volumetric, sofka2017fully}.
We demonstrate this framework in the context of affine and spline-based registration of 3D multi-modal brain MR scans. 

This work is an extension of our conference publication~\citep{yu2022keymorph}, which presented KeyMorph as a robust, unsupervised, affine registration method that aligns learned keypoints.
In this work, we extend our previous submission to parametric, spline-based non-linear transformations with user-specified hyperparameters, and explore the controllability that this affords.
Additionally, we provide a comprehensive treatment of both supervised and unsupervised variants, introduce new training strategies for unsupervised variants, and empirically demonstrate improved results.

\section{Related Work}
\bpara{Classical Methods.}
Pairwise iterative, optimization-based approaches have been extensively studied in medical image registration~\citep{hill2001medical,oliveira2014medical}. These methods employ a variety of similarity functions, types of deformation, transformation constraints or regularization strategies, and optimization techniques. 
% The similarity function quantifies how well a transformed moving image matches a fixed image (or template). 
Intensity-based similarity criteria are most often used, such as mean-squared error (MSE) or normalized cross correlation for registering images of the same modality~\citep{avants2009advanced, avants2008symmetric, hermosillo2002variational}. For registering image pairs from different modalities, statistical measures like mutual information or contrast-invariant features like MIND are popular~\citep{heinrich2012mind,hermosillo2002variational, hoffmann2020learning, mattes2003pet, viola1997alignment}.

% A wide range of transformation models have been used for image registration. Parametric transformations such as rigid, affine, and B-splines have relatively lower computational complexity due to the smaller number of parameters that have to be computed~\cite{avants2009advanced, schnabel2001generic}. The parameters are then often found using gradient-based optimization tools~\cite{modersitzki2004numerical}.
%Deformable transformations are often regularized, with common constraints includes diffusive regularisation, curvature regularization, elastic model, fluid model, topological preservation and diffeomorphism~\cite{ashburner2007fast, bajcsy1989multiresolution, beg2005computing, bro1996fast, fischer2003curvature, staring2007rigidity, thirion1998image, vercauteren2009diffeomorphic}. 

Another registration paradigm first detects features or keypoints in the images, and then establishes their correspondence. This approach often involves handcrafted features~\citep{tuytelaars2008local}, features extracted from curvature of contours~\citep{rosenfeld1971edge}, image intensity~\citep{forstner1987fast, harris1988combined}, color information~\citep{montesinos1998differential, van2005boosting}, or segmented regions~\citep{matas2004robust, wachinger2018keypoint}. Features can be also obtained so that they are invariant to viewpoints~\citep{bay2006surf, brown2005multi,lowe2004distinctive, toews2013feature}. These algorithms then optimize similarity functions based on these features over the space of transformations~\citep{chui2003new,hill2001medical}. This strategy is sensitive to the quality of the keypoints and often suffer in the presence of substantial contrast and/or color variation~\citep{verdie2015tilde}. 

\bpara{Learning-based Methods.}
% More recently, learning-based strategies for image registration have become the dominant approach. 
In learning-based image registration, supervision can be provided through ground-truth transformations, either synthesized or computed by classical methods~\citep{cao2018deformable, dosovitskiy2015flownet, eppenhof2018pulmonary, lee2019image, uzunova2017training, yang2017quicksilver}. Unsupervised strategies use loss functions similar to those employed in classical methods~\citep{balakrishnan2019voxelmorph, dalca2019unsupervised, de2019deep, fan2018adversarial, krebs2019learning, qin2019, wu2015scalable, hoopes2021hypermorph}. Weakly supervised models employ (additional) landmarks or labels to guide training~\citep{balakrishnan2019voxelmorph, fan2019birnet, hu2018label, hu2018weakly}.

% Learning-based approaches achieve excellent results, but have mostly been explored in registration with limited misalignment between the images. In particular, they have been not used for the registration of images that exhibit large (e.g. affine) misalignments, for instance due to major orientation or viewpoint differences. Virtually all prior learning-based registration techniques assume that the images are roughly in the same orientation and position, even when an affine registration component is included. As we show in our experiments, state-of-the-art baselines often fail in the large misalignment scenario. In contrast to these approaches, we view the problem as matched keypoint detection, which can be trained in the absence of supervision, affords robust registration, and yields insights into what parts of the image drive the registration.
%Furthermore, existing deep learning based methods often directly output the spatial transformations (e.g., affine parameters)~\cite{balakrishnan2019voxelmorph, de2019deep, mocanu2021flowreg}, which makes these models hard to interpret. 

Recent learning-based methods compute image features or keypoints~\citep{ma2021image} that can be used for image recognition, retrieval, or registration. 
Learning useful features or keypoints can be done with supervision~\citep{verdie2015tilde,yi2016lift,yi2018learning}, self-supervision~\citep{detone2018superpoint,liu2021same} or without supervision~\citep{barroso2019key, lenc2016learning, ono2018lf}. 
Finding correspondences between pairs of images usually involves identifying the learned features which are most similar between the pair.
In contrast, our method uses a network which extract/generates keypoints directly from the image. 
The keypoints between the moving and fixed image are corresponding (i.e., matched) by construction, and we optimize these corresponding keypoints directly for the registration task (and not using any intermediate keypoint supervision).
% In contrast, our focus is on robust image registration via corresponding keypoints; these prior works aim explicitly to obtain keypoints that are repeatable under different viewpoints and/or image acquisition conditions. Nevertheless, we build on previous ideas and introduce a framework that outputs matched keypoints in 3D space regardless of the initial position of the image. %In addition, these works have exclusively dealt with 2D natural images. 

Learning-based methods for multi-modal registration are of great practical utility and often-studied in the literature.
Most works require, in addition to the moving and fixed image, a corresponding image in a standard space which can be compared and which drives the alignment, usually in the form of segmentations.
\cite{zhang2022twostep} address multi-modal retinal images and handle multi-modality by transforming each image to a standard grayscale image via vessel segmentation. 
A standard feature detection and description procedure is used to find correspondences from these standard images.
Other works \cite{song2022crossmodal} rely on segmentations from ultrasound and magnetic resonance images to align them.
Obtaining these segmentations may be costly, add additional computational complexity to the registration procedure, or be specific to the anatomies/modalities in question.
In contrast, our method can be applied generally to any registration problem.
In addition, we present a variant of our model which only relies on the images themselves during training.
In our experiments, we find that this variant outperforms state-of-the-art baselines while also performing comparably to a variant of our model which leverages segmentations.

\bpara{Hyperparameter-Agnostic Methods.}
A recent line of work focuses on making the model agnostic to the hyperparameter which weights the smoothness prior~\citep{hoopes2021hypermorph,mok2021conditional,wang2022computing}.
These models accept the image pair and hyperparameter as input and modulate the network according to the hyperparameter, either by passing the hyperparameter as input to a hypernetwork which outputs the weights of the main registration network~\citep{hoopes2021hypermorph,wang2022computing}, or by performing instance normalization on the intermediate feature maps~\citep{mok2021conditional}.
% However, these methods are limited by the inherent trade-off between similarity and smoothness that the hyperparameter balances.
% A registration from a high hyperparameter value will be smooth at the cost of similarity, and vice versa. 
% In this work, we propose a strategy which guarantees optimal registrations across every hyperparameter by construction, given a set of learned keypoints.
In this work, we propose an alternative strategy to efficiently produce hyperparameter-specific registration results directly from learned keypoints. In particular, our approach does not require a hypernetwork architecture or any modulation of the architecture itself. 

\section{Differentiable, Closed-Form Coordinate Transformations}
\label{sec:transformations}
\bpara{Notation:} In the following sections, column vectors are lower-case bolded and matrices are upper-case bolded. 
$D$-dimensional coordinates are represented as column vectors, i.e. $\bm{p} \in \mathbb{R}^D$. $D$ is typically 2 or 3. 
$\tilde{\bm{p}}$ denotes $\bm{p}$ in homogeneous coordinates, i.e. $\tilde{\bm{p}} = [\bm{p}, 1]^T$.
Superscripts in parentheses $\bm{p}^{(i)}$ index over separate instances of $\bm{p}$ (e.g. in a dataset), whereas subscripts $\bm{p}_i$ denotes the $i$'th element of $\bm{p}$.

We present two parametric transformation families that can be derived in closed-form, from corresponding keypoint pairs.
Suppose we have a set of $N$ corresponding keypoint pairs $\{(\bm{p}^{(i)}, \bm{q}^{(i)})\}_{i=1}^N$, where $\bm{p}^{(i)}, \bm{q}^{(i)} \in \mathbb{R}^D$ and $N > D$.
For convenience, let $\bm{P} := \begin{bmatrix}\bm{p}^{(1)} & ...  & \bm{p}^{(N)} \end{bmatrix}\in \mathbb{R}^{D\times N}$, and similarly for $\tilde{\bm{P}}$ and $\bm{Q}$. 
Define $\mathcal{T}_\theta : \mathbb{R}^D \rightarrow \mathbb{R}^D$ as a family of coordinate transformations, where $\theta \in \Theta$ are parameters of the transformation.
% We seek a solution to the following optimization problem:
% \begin{equation}
% \opttran = \argmin_{\theta} \sum_{i=1}^N \left(\mathcal{T}_\theta(\bm{p}^{(i)})-\bm{q}^{(i)}\right)^2.
% \end{equation}
% where $\mathcal{T}_\theta(\bm{P})$ is short-hand for applying the transformation $\mathcal{T}_\theta$ to each column of $\bm{P}$.
% We seek a function $\opttran : \mathbb{R}^{D\times N} \times \mathbb{R}^{D\times N}  \rightarrow \Theta$ which outputs the optimal parameters for the chosen transformation family given corresponding sets of keypoints.

\subsection{Affine}
Affine transformations are represented as a matrix multiplication of $\bm{A} \in \mathbb{R}^{D \times (D+1)}$ with a coordinate in homogeneous form:
\begin{equation}
    \mathcal{T}_\theta(\bm{p}) = \bm{A}\tilde{\bm{p}},
\end{equation}
where the parameter set is the elements of the matrix,~\mbox{$\theta = \{\bm{A}\}$}.

Given $N$ corresponding keypoint pairs, there exists a differentiable, closed-form expression for an affine transformation that aligns the keypoints:
\begin{align} 
\opttran(\bm{P}, \bm{Q}) &:= \argmin_{\theta} \sum_{i=1}^N \left(\bm{A} \tilde{\bm{p}}^{(i)} - \bm{q}^{(i)}\right)^2 \\
    %&= \argmin_{\theta} \norm{\bm{A} \tilde{\bm{P}}-\bm{Q}}_F \\
    &= \bm{Q}\tilde{\bm{P}}^{T}(\tilde{\bm{P}}\tilde{\bm{P}}^{T})^{-1}.
\label{eq:closeform}
\end{align}
%Note that solving for $\theta^*$ is a differentiable operation.
%
% We provide the derivation in Appendix~\ref{appendix:proof}.

A derivation is provided in the Appendix. This is the least-squares solution to an overdetermined system, and thus in practice the points will not be exactly matched due to the restrictive nature of the affine transformation.
This restrictiveness may be alleviated or removed by choosing a transformation family with additional degrees of freedom, as we detail next.

\subsection{Thin-Plate Spline}
The application of the thin-plate spline (TPS) interpolant to modeling coordinate transformations yields a parametric, non-rigid deformation model which admits a closed-form expression for the solution that interpolates a set of corresponding keypoints~\citep{bookstein1989tps,donato2002approximatetps,rohr2001landmark}.
This provides additional degrees of freedom over the affine family of transformations, while also subsuming it as a special case. 

For the $d$'th dimension, the TPS interpolant $\mathcal{T}_{\theta_d} : \mathbb{R}^D \rightarrow \mathbb{R}$ takes the following form:
\begin{equation}
    \mathcal{T}_{\theta_d}(\bm{p}) = (\bm{a}_d)^T \tilde{\bm{p}} + \sum_{i=1}^N w_{i,d} U\left(\norm{\bm{p}^{(i)} - \bm{p}}_2\right),
\end{equation}
where $\bm{a}_d \in \mathbb{R}^{D+1}$ and $\{w_{i,d}\}$ constitute the transformation parameters $\theta_d$ and $U(r) = r^2 \ln(r)$. 
We define $\bm{A} \in \mathbb{R}^{(D+1) \times D}$ and $\bm{W} \in \mathbb{R}^{N \times D}$ as the collection of all the parameters for $d=1, ..., D$.
Then, the parameter set is $\theta = \{\bm{A}, \bm{W}\}$.

% Let us now consider the following optimization problem: 
% \begin{equation}
%     \argmin_{\theta'} \sum_{i=1}^N \left(T_{\theta'}\left(\bm{p}^{(i)}\right) - \bm{q}_d^{(i)}\right)^2 + \lambda I_{T},
% \end{equation}
% where $\lambda >0$ is a hyper-parameter,
% $I_{T}$ is the \textit{bending energy}:
% %
% \begin{equation}
%     I_{T} = \int_{\mathbb{R}^D} \norm{\nabla^2 T}_F^2 d\bm{p}_1...d\bm{p}_D,
% \end{equation}
% %
% and $\nabla^2 T$ is the matrix of second-order partial derivatives of $T$.

This form of $\mathcal{T}$ minimizes the \textit{bending energy}:
\begin{equation}
    I_{\mathcal{T}} = \int_{\mathbb{R}^D} \norm{\nabla^2 \mathcal{T}}_F^2 d\bm{p}_1...d\bm{p}_D,
\end{equation}
where~$\norm{\cdot}_F$ is the Frobenius norm and $\nabla^2 \mathcal{T}$ is the matrix of second-order partial derivatives of $\mathcal{T}$. 
% Here, $\theta' = \{\bm{a}, w_1, ..., w_N\}$. 
% Define $\bm{A} \in \mathbb{R}^{(D+1) \times D}$ and $\bm{W} \in \mathbb{R}^{N \times D}$ as the collection of all the parameters for $d=1, ..., D$.
% Then, the parameter set is $\theta = \{\bm{A}, \bm{W}\}$.
For each $\theta_d$, we impose interpolation conditions $\mathcal{T}_{\theta_d}(\bm{p}^{(i)}) = \bm{q}_d^{(i)}$ and enforce $\mathcal{T}$ to have square-integrable second derivatives: 
\begin{equation}
    \sum_{i=1}^N w_{i,d} = 0 \ \ \text{and} \ \ 
    \sum_{i=1}^N w_{i,d} \bm{p}_d = 0 \ \ \forall d \in \{1, ..., D\}.
\end{equation}
Given these conditions, the following system of linear equations solves for $\theta$: 
\begin{equation}
    \bm{\Psi} \theta :=
    \begin{bmatrix}
    \bm{K} & \bm{L} \\
    \bm{L}^T & \bm{O}
    \end{bmatrix}
    \begin{bmatrix}
    \bm{W} \\
    \bm{A}
    \end{bmatrix}
    =
    \begin{bmatrix}
    \bm{Q}^T \\
    \bm{O}
    \end{bmatrix}
    := \bm{Z}.
    \label{eq:block-tps}
\end{equation}
Here, $\bm{K} \in \mathbb{R}^{N\times N}$ where $\bm{K}_{ij} = U\left(\norm{\bm{p}^{(i)} - \bm{p}^{(j)}}_2\right)$, $\bm{L} \in \mathbb{R}^{N \times (D+1)}$ where the $i$'th row is $(\tilde{\bm{p}}^{(i)})^T$, and $\bm{O}$ is a matrix of zeros with appropriate dimensions.
Thus,
\begin{equation}
    \opttran(\bm{P}, \bm{Q}) := \bm{\Psi}^{-1} \bm{Z}.
\end{equation}
Solving for $\theta^*$ is a differentiable operation.

The interpolation conditions can be relaxed (e.g. under the presence of noise) by introducing a regularization term:
\begin{equation}
    \argmin_{\theta_d} \sum_{i=1}^N \left(\mathcal{T}_{\theta_d}\left(\bm{p}^{(i)}\right) - \bm{q}_d^{(i)}\right)^2 + \lambda I_{\mathcal{T}}
\end{equation}
where $\lambda>0$ is a hyperparameter that controls the strength of regularization.
As $\lambda$ approaches $\infty$, the optimal $\mathcal{T}$ approaches the affine case (i.e. zero bending energy). 
This formulation can be solved exactly by replacing $\bm{K}$ with $\bm{K}+\lambda \bm{I}$ in Eq.~\eqref{eq:block-tps}.
Importantly, $\theta$ and the optimal $\theta^*(\bm{P}, \bm{Q})$ exhibits a dependence on $\lambda$.

\section{KeyMorph}
Let~$(\moveimg, \fiximg)$ be moving (source) and fixed (target) image\footnote{Although we consider 3D volumes in this work,~\modelname~is agnostic to the number of dimensions. The terms ``image'' and ``volume''  are used interchangeably.} pairs, possibly of different contrasts or modalities.  
Additionally, we denote by  $\mathcal{T}_{\theta}$ a parametric coordinate transformation with parameters $\theta$, such as those discussed in Section~\ref{sec:transformations}. 
Our goal is to find the optimal transformation $\mathcal{T}_{\theta^*}$ such that the registered image~$\alignimg = \moveimg \circ \mathcal{T}_{\theta^*}$ aligns with the fixed image~$\fiximg$, where~$\circ$ denotes the spatial transformation of an image.

\modelname~derives the optimal $\theta^*$ by detecting $N$ \textit{corresponding keypoints} $\bm{P}, \bm{Q} \in \mathbb{R}^{D\times N}$ from $\moveimg$ and $\fiximg$, respectively.
Furthermore, by choosing a transformation family whose optimal parameters admits a closed-form and differentiable solution $\opttran(\bm{Q}, \bm{P})$ as a function of these keypoints, the keypoints themselves are learned by a neural network in an end-to-end fashion\footnote{Note that the arguments of $\theta^*$ are switched (i.e. the transformation $\mathcal{T}$ is applied to the ``fixed" keypoints $\bm{Q}$), because we are seeking a transformation that takes us from fixed image coordinates to moving image coordinates in order to resample the moving image.}.

We define a CNN as $\nn$ with parameters $\nnparam$, which detects $N$ corresponding keypoints from both the moving and fixed image: $\bm{P} = \nn(\moveimg)$ and~$\bm{Q} = \nn(\fiximg)$. 
Supposing we have a dataset of such pairs, the general~\modelname~objective is:
\begin{align}
\begin{split}
\argmin_\nnparam \ &\mathbb{E}_{(\moveimg, \fiximg)} \  \mathcal{L}_{sim}\left(\moveimg \circ \mathcal{T}_{\theta^*}, \fiximg\right)  \\
&\text{where } \ \ \theta^* = \opttran\left(\nn(\fiximg), \nn(\moveimg) \right)
\label{eq:objective}
\end{split}
\end{align}
where $\mathcal{L}_{sim}(\cdot, \cdot)$ measures image similarity between its two inputs.
Fig.~\ref{fig:proposed_model} depicts the proposed model architecture.
The closed-form optimal solution $\theta^*$ can depend on a hyperparameter $\lambda$, such as in TPS, which can be set to a constant or sampled from a distribution $\lambda \sim p(\lambda)$ during training.
Once the model is trained, at test (inference) time, one can swap the transformation with a different model (e.g, replace affine with TPS) or use a different hyperparameter value, which would yield a different alignment based on the same keypoints. 

There are several advantages to our formulation over prior learning-based methods.
1) The closed-form solution is not sensitive to the relative placement of the keypoints. Thus, robustness to large misalignments is achieved if the CNN is equivariant with respect to input image deformation. 
Intuitively, this is an easier task for the network to learn as compared to having to learn both correspondences and the transformation directly from the image pairs.
% sensitive to the relative placement of the keypoints.
2) Visualizing the keypoints enables the user to interpret what parts of the image is driving the alignment.
In contrast to deformation fields which encode correspondence and transformation in a dense velocity space, keypoints are easily visualized and overlaid on the image space. Furthermore, the keypoints are guaranteed to be anatomically consistent by construction; as a result, they can be used to understand and verify model behavior, or leveraged for downstream tasks.
3) With careful design of the keypoint detection network (see Section~\ref{sec:arch}), translational-equivariance can be achieved.
4) Once the model is trained, different coordinate transformations or different transformation hyperparameters can be used to generate a dense set of registrations at test-time; the controllable nature of this framework enables the user to select the preferred registration.  
Note that these registrations require no modulation of the CNN to produce.
% In this work, we choose $\mathcal{T}$ to be the TPS transformation, where $\lambda$ controls the degree of non-linearity as detailed in Section~\ref{sec:transformations}.

\subsection{Keypoint Detector Architecture}
\label{sec:arch}
\modelname~is a framework that can leverage any deep learning-based keypoint detector~\citep{ma2021image,detone2018superpoint,barroso2019key}.
In this work, we are interested in preserving translation equivariance; to this end, we leverage a center-of-mass (CoM) layer~\citep{ma2020volumetric, sofka2017fully} as the final layer, which computes
the center-of-mass for each of the $N$ activation maps. 
This specialized layer is (approximately) translationally-equivariant and enables precise localization. 
Our architecture backbone consists of convolutional layers followed by instance normalization~\citep{ulyanov2016instance}, ReLU activation, and 2x downsampling via strided convolution, as shown in Fig. \ref{fig:proposed_model}. 
% We provide more details and compare CoM to fully-connected layers in Appendix~\ref{appendix:com}.

\subsection{Training and Model Variants}
\label{sec:variants}
We present two training strategies corresponding to two model variants of~\modelname.
For both variants, we apply random affine transformations to the moving image $\bm{x}_m$ as an augmentation strategy, the details of which are in Section~\ref{sec:model_and_training_details}.

\subsubsection{Supervised}
In what we refer to as the ``supervised'' setting, the model may exploit segmentations during training. 
In this work, we use soft-Dice for $\mathcal{L}_{sim}$, which measures volume overlap of the moving and registered label maps~\citep{balakrishnan2019voxelmorph, hoffmann2020learning, hu2018label}.
Note that label maps are only used to compute the loss and are not an input to the model.

\subsubsection{Unsupervised}
In what we refer to as the ``unsupervised'' setting, we assume we have access to intensity images only. 
Here, $\mathcal{L}_{sim}$ is mean squared error (or MSE) on pixel intensity values and is computed only on \textit{same-modality} image pairs.

To ensure that keypoints remain consistent across modalities, we add an additional loss that penalizes within-subject multi-modal (e.g. T1, T2, PD-weighted MRI) keypoint deviation.
Thus, in the unsupervised setting, we alternate between mini-batches containing multi-subject uni-modal image pairs (where pixel MSE is the loss function as in Eq.\eqref{eq:objective}) and the following same-subject multi-modal image pairs (where keypoint MSE is the loss function): 
\begin{equation}
\mathcal{L}_u(\nnparam) = \mathbb{E}_{(\bm{x}_1, \bm{x}_2)}\norm{\nn(\bm{x}_1) - \nn(\bm{x}_2)}_F,
    \label{eq:keypoint_loss}
\end{equation}
where $(\bm{x}_1, \bm{x}_2)$ is a pair of images from the same subject with different modality.
This is equivalent to minimizing the sum of the losses with equal weight, which we found to work well in practice.
We apply random affine transformations to the same-subject multi-modal image pairs as an augmentation strategy.

Note that, contrary to other multi-modal registration methods, $\mathcal{L}_{sim}$ is set to MSE for both uni-modal and multi-modal training.

% We assume that we have access to aligned multi-modal  volumes for each subject.
% Suppose we have a dataset composed of same subject, different modality image pairs $(\bm{x}_1, \bm{x}_2)$.
% Then, the loss we minimize is:
% %
% \begin{equation}
%     \mathcal{L}_u(\nnparam) = \mathbb{E}_{(\bm{x}_1, \bm{x}_2)}\norm{\nn(\bm{x}_1) - \nn(\bm{x}_2)}_F,
%     \label{eq:keypoint_loss}
% \end{equation}
% We apply identical random affine deformations to $\bm{x}_1$ and $\bm{x}_2$ as an augmentation strategy.
% The final objective is:
% %
% \begin{equation}
%     \argmin_\nnparam \mathcal{L}(\nnparam) + \alpha\mathcal{L}_u(\nnparam) 
% \end{equation}
% %
% where $\alpha>0$ is a hyperparameter chosen by cross-validation.

%
\begin{figure*}[t]
\centering
\includegraphics[width=\textwidth]{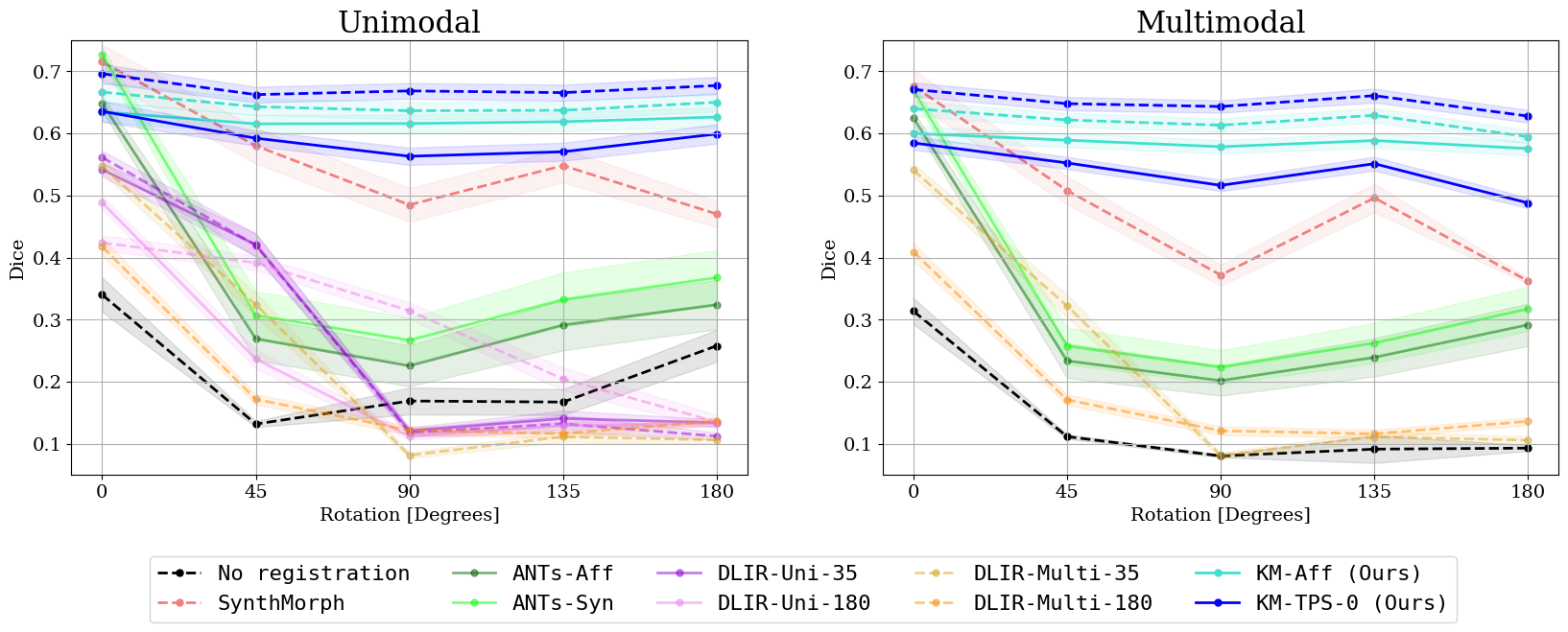}
   \caption{\modelname~and baseline performance across varying rotation angle. The Dice score is averaged for all test subjects and brain anatomical regions. Solid lines denote unsupervised models and dashed lines denote supervised models (i.e. with or without ground-truth segmentations). Shaded regions denote standard deviation.}
%   See Section 4.3 for details on the naming scheme.
\label{fig:box_plot_over_augs}
\end{figure*}
\subsection{Self-supervised Pretraining}
We employ the following self-supervised pre-training strategy to aid in keypoint detector initialization. 
Consider a single subject and its set of aligned different-modality scans \{$\bm{x}^{(i)}\}$.
We pick a random set of keypoints~$\bm{P}_0$ by sampling uniformly over the image coordinate grid. 
In each mini-batch, we apply random affine and nonlinear transformations to~$\bm{x}^{(i)}$ and~$\bm{P}_0$, and minimize the following keypoint loss:
\begin{equation}
\argmin_\nnparam \sum_i \mathbb{E}_{\bm{A}, \phi}\norm{\phi(\bm{A}\bm{P}_0) - \nn\left(\bm{x}^{(i)} \circ \bm{A} \circ \phi\right)}^2_2. 
\end{equation} 
Here, $\bm{A}$ is an affine transformation drawn from a uniform distribution over the parameter space. $\phi$ is a nonlinear transformation generated by integrating a random stationary velocity field (see~\cite{ashburner2007diffeomorphic,dalca2018diffeomorphic,hoffmann2020learning} for more details).

Empirically, we find that this relatively lightweight pretraining step is necessary to encourage the keypoints to spread out across the image. 
Without it, the training dynamics suffer due to being stuck in a local minimum where the keypoints cluster tightly in the middle of the image, leading to suboptimal solutions.

% In the supervised setting, ground-truth keypoints may be generated by taking the center-of-mass of segmentation labels for each $\bm{x}$ in the dataset, thus eliminating the restriction of pretraining on a single subject. 
% However, this restricts the number of keypoints to be the number of distinct labels.
% Empirically, we find that the single-subject pretraining performs well while providing the flexibility to choose the number of keypoints.
%
% This pretraining step imbibes the keypoint detector network with invariance to affine deformations as well as modality. 
% We found that \modelname~learns to detect anatomically-consistent keypoints \textit{across modalities}, even though it is trained on same-modality pairs.
% In our experiments, we demonstrate how this can be used to perform multi-modal registration at test-time.

%
\begin{figure*}[t]
\centering
\includegraphics[width=\textwidth]{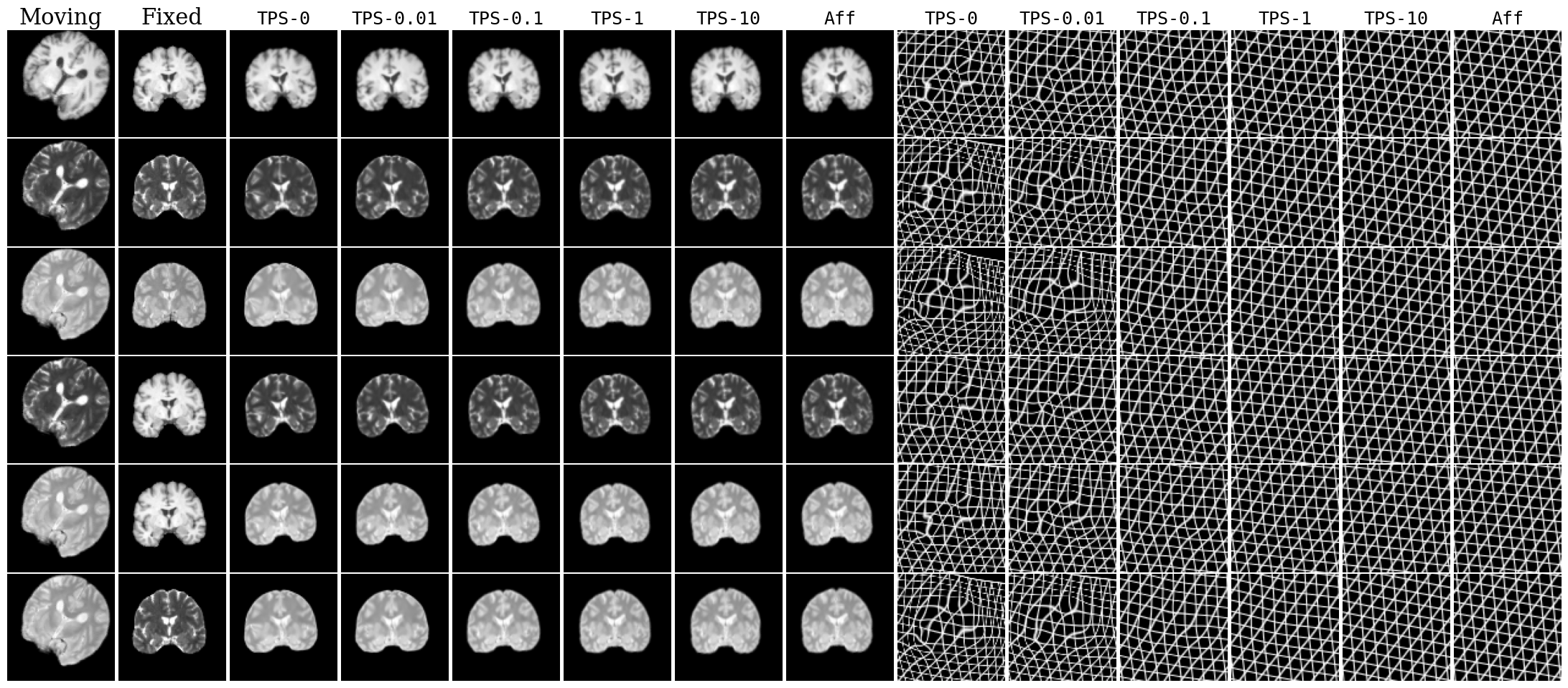}
   \caption{Representative registration results on different volumes from a pair of subjects, for unsupervised \texttt{KM} variants. Registered images are performed for different TPS $\lambda$'s and affine transformations. 
   Deformed grids are depicted.}
\label{fig:warp_grid}
\end{figure*}
\section{Experimental Setup}
\begin{table}[t]
\centering
\begin{tabular}{lcc}
\hline
\textit{Model} & \textit{CPU Time (s)} & \textit{GPU Time (s)} \\ \hline 
\texttt{ANTs-Aff}           & 50.59$\pm$1.18            & -                         \\
\texttt{ANTs-Syn}           & 52.70$\pm$0.72                      & -                         \\
\texttt{DLIR}               & 1.49$\pm$0.09             & 0.02$\pm$0.001            \\
\texttt{KM-Aff} (Ours)      & 2.61$\pm$0.53             & 0.09$\pm$0.001            \\ 
\texttt{KM-TPS} (Ours)      & 4.29$\pm$0.17             & 0.23$\pm$0.009            \\ \hline
\end{tabular}%
\caption{Average computation time across different models during testing.}
\label{tab:timing}
\end{table}

\label{sec:experiments}

%<summary> 

\subsection{Dataset} 
We experiment on the IXI brain MRI dataset\footnote{\url{https://brain-development.org/ixi-dataset/}}. Each subject has T1, T2, and PD-weighted 3D MRI scans in spatial alignment. 
We perform the following standard preprocessing steps: resampling to 1mm isotropic, rescaling intensity values between $[0, 1]$, and padding to $256^3$ image size.
We partition the 577 total subjects into sets of 427, 50, and 100 for training, validation, and testing, respectively. 
We perform standard skull stripping~\citep{kleesiek2016deep} on all images. 

We use a pre-trained and validated SynthSeg model~\citep{billot2020learning} to automatically delineate 23 regions of interest (ROIs)\footnote{ROIs were pallidum, amygdala, caudate, cerebral cortex, hippocampus, thalamus, putamen, white matter, cerebellar cortex, ventricle, cerebral white matter, and brainstem.}. 
These segmentations were used for training a subset of (supervised) models, as described below. 
Furthermore, all performance evaluations were based on examining the overlap of ROIs in the test images.

\subsubsection{Test-time Performance Evaluation}
We use each testing subject as a moving volume~$\moveimg$, paired with another random test subject treated as a fixed volume~$\fiximg$. 
For all test volumes, we use 23-label segmentation maps to quantify alignment. 
We simulate different degrees of misalignment by transforming~$\moveimg$ using rotation. 
1 to 3 axes are chosen randomly and a uniform random rotation is applied up to a given degree. 
We use the predicted transformation to resample the moved segmentation labels on the fixed image grid. 
We quantify alignment quality and properties of the transformation using Dice overlap score, Hausdorff distance, standard deviation of the Jacobian determinant, and percentage of voxels less than 0 in the Jacobian determinant.

We perform registration across all combinations of available modalities (registering T1 to T1, T1 to T2, etc). All pairings and amount of transformations are kept the same across the registration experiments. 

\subsection{Baselines}
\label{sec:baselines}
% We compare~\modelname~against affine and nonlinear variants of a classical and learning-based method. 
% Since we experiment under scenarios of (possibly) large initial misalignment, we do not compare against methods which require an initial robust registration step to a fixed template, e.g.~\cite{hoffmann2020learning,hoopes2021hypermorph}.

\textbf{Advanced Normalizing Tools (ANTs)} is a widely-used software package which is state-of-the-art for medical image registration~\citep{avants2009advanced}. For the affine model, we use ``TRSAA'' affine implementation (\texttt{ANTs-Aff}), which consist of translation, rigid, similarity and two affine transformation steps. The volumes are registered successively at three different resolutions: 0.25x, 0.5x and finally at full resolution. At 0.25x and 0.5x resolution, Gaussian smoothing with~$\sigma$ of two and one voxels is applied, respectively. 
For non-linear registration, we use ``SyNRA'' (\texttt{ANTs-Syn}), which consists of an initial rigid and affine alignment step followed by Symmetric Normalization~\citep{avants2008syn}. 
For both \texttt{ANTs-Aff} and \texttt{ANTs-Syn}, we use the ANTs affine initializer at large deformation, which conducts a grid search over a range of rotations and translations to help find a good initialization. Finally, we used mutual information as the similarity metric for both models, which is suitable for registering images with different contrasts.

\textbf{Deep Learning for Image Registration (DLIR)} is a recent learning-based method that supports affine and deformable image registration~\citep{de2019deep}, and which we believe is representative of many deep learning models which leverage a Spatial Transformer Network~\citep{jaderberg2015spatial} (STN). 
STNs were originally used to improve class prediction accuracy and has since become the backbone of many subsequent works in image registration~\citep{de2017end, lee2019imageandspatial, balakrishnan2019voxelmorph}. %using a CNN to taking fixed and moving image to a CNN, which then outputs a spatial transformation. This transformation is then applied to a sampling grid which is used to produce a registered image. STNs are trained end-to-end using a similarity metric.
%
%The type of transformation these models learn depends on the output of the neural network. 
% We implemented a STN to serve as a deep learning baseline. 
For a direct comparison, we used the same backbone architecture as \modelname~replacing the center-of-mass layer with a fully-connected (FC) layer which outputs 12 parameters for the 3D affine transformation.\footnote{
% For STN, the \texttt{Tanh} layer was removed at the end. 
We used instance norms for multimodal DLIR and batch norms for unimodal DLIR, which we found to work well in practice.} %In Appendix \ref{appendix:com}, we also investigated a \modelname~implementation with the same FC layer as DLIR. 

For all DLIR training, we use the following ranges for random augmentations: translations $[-15, 15]$ voxels, scaling factor $[0.8, 1.2]$, and shear $[-0.1, 0.1]$.
We find that DLIR often cannot register image pairs with large misalignments, especially under large rotation misalignment. 
We alleviate this by using more aggressive rotation augmentation during training. 
We consider two different amounts of rotation for the training of DLIR: maximum~$\pm 35 \degree$ or~$\pm 180 \degree$. We use the same loss function and training scheme as we used for \modelname. 
% We found that this approach improved the performance of DLIR for large transformations.
We trained separate DLIR models for each modality as it produces better results than training DLIR across modalities with mutual information. We also trained \textit{supervised} modality-specific and multi-modal DLIR models using a soft-Dice loss computed on the aligned segmentation maps~\citep{lee2019image}. 
%Using soft-Dice also enabled a multi-modal DLIR model that can accept any modality image pair as input.

Altogether, we implement six different DLIR variants. 
The naming scheme follows the convention \texttt{DLIR-<mod>-<degree>}, where \texttt{<mod>} denotes whether the model was trained on a single (\texttt{Uni}) or multiple (\texttt{Multi}) modalities and \texttt{<degree>} denotes the maximum angle of rotation for augmentation. We train supervised and unsupervised versions of unimodal DLIR, and only train supervised versions of multimodal DLIR.

\textbf{SynthMorph} is a recently-proposed deep learning model which achieves agnosticism to modality/contrast by leveraging a generative strategy for synthesizing diverse images, thereby supporting multi-modal registration~\citep{hoffmann2022synthmorph}.
Like DLIR, it accepts as input the moving and fixed images but outputs a dense deformation field instead of global affine parameters, which is a common strategy in many well-performing registration models~\citep{balakrishnan2019voxelmorph}.
However, a crucial assumption of many of these models is that the image pair must first be affine registered.
Although \modelname~does not have this assumption, we follow this required step to maximize the performance of SynthMorph.
Following their suggested pre-registration method, we first affine-register each image pair using \texttt{mri-robust-register}~\footnote{\url{https://surfer.nmr.mgh.harvard.edu/fswiki/mri_robust_register}} from the Freesurfer package~\citep{fischl2012freesurfer}.

\subsection{Model and Training Details}
\label{sec:model_and_training_details}
We train six variants of \modelname~(\texttt{KM}) corresponding to six \textit{pre-determined} transformations: 1 model using an affine transformation and 5 models using a TPS transformation with $\lambda \in \{0, 0.01, 0.1, 1, 10\}$.
We refer to these models as \texttt{KM-Aff} and \texttt{KM-TPS-$\lambda$}.
Additionally, we train a variant using a TPS transformation where $\lambda$ is sampled \textit{stochastically} during training from a log-uniform distribution $p(\lambda) = LogUnif(0, 10)$.
We refer to this model as \texttt{KM-LogUnif}.
We used $N=128$ keypoints throughout our experiments, and analyze the effect of changing this number in Section~\ref{sec:experiment3}.

For all models, we used a batch size of 1 image pair and the Adam optimizer~\citep{kingma2017adam} for training.
The following uniformly-sampled augmentations were applied to the moving image across all dimensions during training: rotations $[-180\degree, +180\degree]$, translations $[-15, 15]$ voxels, scaling factor $[0.8, 1.2]$, and shear $[-0.1, 0.1]$.
Note that this is the same augmentation strategy as DLIR variants with $[-180\degree, +180\degree]$ rotation augmentation.
All training and GPU testing was performed on a machine equipped with an Intel Xeon Gold 6126 processor and an Nvidia Titan RTX GPU. CPU testing was performed on a machine equipped with an AMD EPYC 7642 48-Core Processor. All models were implemented in Pytorch. 

\section{Results}
\subsection{Main Results}
\subsubsection{\modelname~is robust to large misalignments}
We analyze the performance of baselines and our proposed \modelname~under conditions of large initial misalignments in terms of rotation.
Fig.~\ref{fig:box_plot_over_augs} plots overall Dice across rotation angle of the moving image for baselines and two variants of \modelname: \texttt{KM-Aff} and \texttt{KM-TPS-0}. 

We find that all DLIR models suffer substantially as the rotation angle increases. Training with aggressive augmentation increases performance for test pairs with large misalignments, but reduces the accuracy for those with smaller misalignments. %Some capacity of the neural networks is spent in learning the large deformation.
 %However, learning the affine parameters directly is very difficult and the accuracy of the smaller angles suffers as a result. 
Using supervision (dashed \texttt{DLIR} lines) leads to improved accuracy. 
For unimodal registration, the \texttt{DLIR} model that was trained with all modalities (\texttt{DLIR-Multi}) did not produce better accuracy than a model that was trained with each modality separately.
\texttt{ANTs} yields excellent results when the initial misalignment is small (e.g. near 0 degrees of rotation). However, the accuracy drops substantially when the misalignment exceeds this range. 

In contrast to these models, \modelname~variants performed well across all types of transformations and ranges, with only marginal drops in accuracy in large misalignments. 
In the case where we have access to ROIs during training, we find that \modelname~trained with Dice outperforms all models except \texttt{ANTs-Syn}. 
However, the unsupervised variant trained with MSE still yields excellent accuracy across all settings and is only minimally suboptimal compared to its supervised counterpart. 

The supervised variant of \texttt{KM-TPS-0} outperforms its \texttt{KM-Aff} counterpart, whereas the opposite is true in unsupervised variants; we attribute this to the increased expressivity of low $\lambda$ in the TPS transformation overfitting to the MSE loss function at the cost of Dice performance (empirically, we observe that low $\lambda$ achieves better performance in terms of MSE compared to high $\lambda$ for unsupervised variants, see Fig.~\ref{fig:keypoint_box}).

We provide qualitative results in Fig.~\ref{fig:warp_grid}, and compare the computational time across different models in Table~\ref{tab:timing}. 
A comprehensive table of all methods and initial misalignments separated by modality are in Table~\ref{tab:big-table}.
Additional metrics including Hausdorff distance and Jacobian-based metrics are presented in Fig.~\ref{fig:hd_J_metrics}.
Qualitative registrations for all models is provided in Fig.~\ref{fig:qualitative-rot0} and \ref{fig:qualitative-rot90}.
Overall, the \modelname~(\texttt{KM}) variants outperform other learning-based baselines, and \modelname~performs comparably or often better (at large misalignments) than the state-of-the-art ANTs registration, while requiring substantially less runtime.

\subsubsection{\modelname~supports multi-modal registration}
\begin{figure}[t]
\centering
\includegraphics[width=\linewidth]{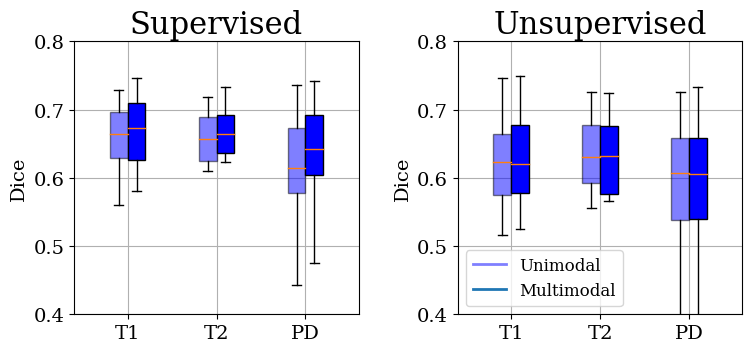}
   \caption{Multimodal KeyMorph (light boxes) vs. unimodal KeyMorph (dark boxes) Dice performance across individual modalities. Left panel shows KeyMorph variants trained with supervised loss. Right panel shows KeyMorph variants trained with unsupervised loss. We observe that our proposed multi-modal training strategy achieves comparable results to the modality-specific models, for both supervised and unsupervised variants.}
\label{fig:unimodal_vs_multimodal_box}
\end{figure}
\begin{figure}[t]
\centering
\includegraphics[width=0.8\linewidth]{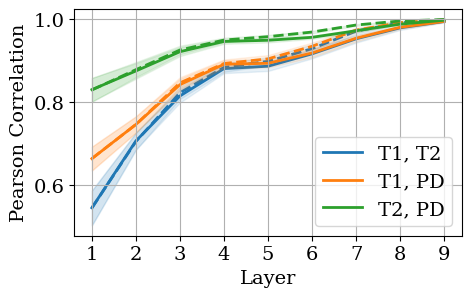}
   \caption{Pearson correlation of feature maps across layers of trained unsupervised (solid) and supervised (dashed) model. Correlation is computed between each combination of (T1, T2, and PD) multi-modal volume pairs, for a fixed subject. Final value is average across subjects, and shaded regions denote standard deviation.}
\label{fig:plot_corr}
\end{figure}
To test the efficacy of the multi-modal training strategy of~\modelname, we compare the performance of a single KeyMorph model trained with our multi-modal strategy against 3 separate KeyMorph models on 3 different modalities. 
Results are shown in Fig.~\ref{fig:unimodal_vs_multimodal_box}, which demonstrate that the multi-modal training strategy achieves comparable results to the modality-specific models.

We hypothesize that in order to remain consistent across modalities, the keypoint detector must learn to be invariant to contrasts. 
To test this, we plot the Pearson correlation of feature maps for multi-modal image pairs, across the layers of the CNN (see Fig.~\ref{fig:plot_corr}).
For a given subject and a given layer, we extract the intermediate feature maps for the 3 modalities and compute the Pearson correlation between every pair of modalities. 
Then, we average across all subjects, for each layer.
We observe that the correlation increases with the depth of the network, indicating that the features become more invariant to the input modality at deeper layers.

\subsubsection{CoM layer outperforms FC layer}
\begin{figure}[t]
\centering
\includegraphics[width=0.90\linewidth]{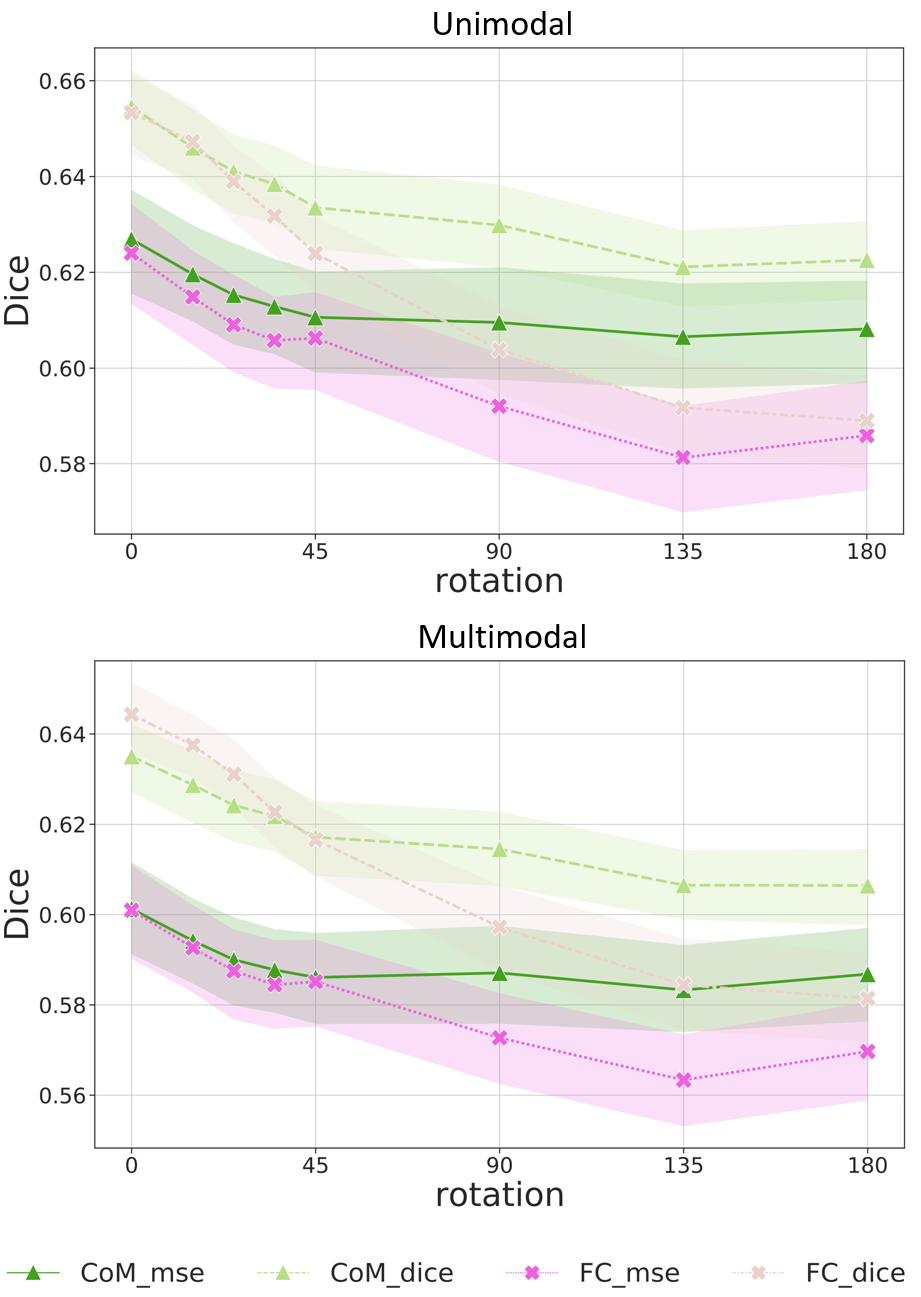}
   \caption{Performance comparison between KeyMorph models that uses center-of-mass
(CoM) and fully connected (FC) layers to predict keypoints. The suffix \texttt{mse} and \texttt{dice} represent the unsupervised and supervised version of KeyMorph, respectively. Shaded regions denote standard deviation.}
\label{fig:fc_vs_com}
\end{figure}
We test whether the translation-equivariance properties of the CoM layer leads to improved registration performance.
Fig.~\ref{fig:fc_vs_com} shows results on KeyMorph variants trained with the CoM layer vs. an FC layer. 
We see that the CoM outperforms FC over the range of transformations.

\subsubsection{Multiple registrations at test-time provide nuanced predictions}
\begin{figure}[t]
\centering
\includegraphics[width=\linewidth]{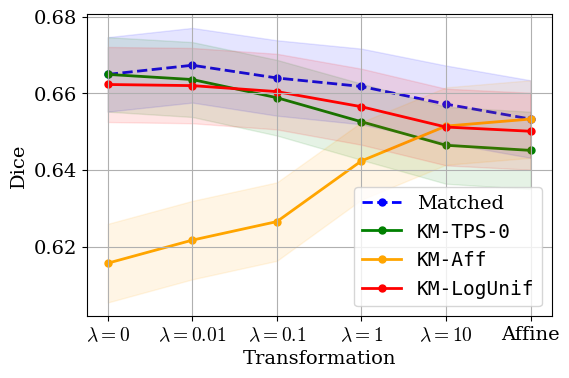}
   \caption{Dice performance over transformation variants. Solid lines are single, pretrained models evaluated on all transformations at test-time. Dashed blue line is multiple, pretrained models on matched transformation at test-time, which is an upper-bound on performance. Shaded regions denote standard deviation.}
\label{fig:robust_transform}
\end{figure}
\begin{figure}[t]
\centering
\includegraphics[width=\linewidth]{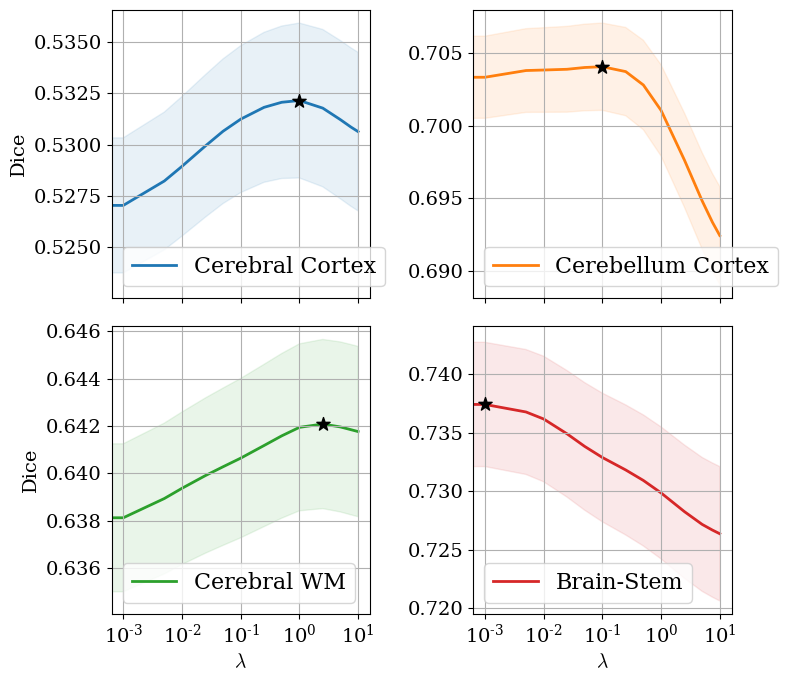}
   \caption{Dice performance over $\lambda$ of \texttt{KM-LogUnif} separated by select ROIs. Stars denote maximum values. Shaded regions denote standard deviation.}
\label{fig:interactive_roi}
\end{figure}
Given learned keypoints, \modelname~can generate a dense set of registrations at test-time corresponding to different transformations.
% In contrast to hyperparameter-agnostic methods~\cite{hoopes2021hypermorph,mok2021conditional}, our method produces optimal registrations given a set of keypoints by construction across the entire hyperparameter space, since the transformation parameters are computed in closed-form. 
It follows that the quality of registrations is subject to whether a single set of learned keypoints can perform well across many different transformations.
% In this experiment, we demonstrate that this is the case, and further demonstrate a downstream utility that multiple registrations can afford.

To this end, we analyze the ability of learned keypoints to generalize to new transformation hyperparameters.
In Fig.~\ref{fig:robust_transform}, each solid line is generated by taking the corresponding model and evaluating it on different transformations at test-time. 
The dashed ``Matched" line denotes the performance of multiple models trained on its corresponding transformation, and is an upper-bound on performance. 
As expected, we notice that \texttt{KM-TPS-0} performs well at low $\lambda$ values and vice versa for \texttt{KM-Aff}.
By comparison, \texttt{KM-LogUnif} performs comparably to these fixed models at the endpoints, and outperforms both near the middle region (i.e. $\lambda \in [0.1, 10]$).
Overall, it performs comparably to the upper-bound across all transformations.
We conclude that sampling from a hyperparameter distribution allows the learned keypoints to be useful for a range of transformations, and that a single trained model can be used in lieu of multiple trained models on single settings of the hyperparameters.

To demonstrate a downstream utility that multiple optimal registrations afford (and following \cite{hoopes2021hypermorph}), we study the relationship between regional Dice scores and the hyperparameter value. 
Fig.~\ref{fig:interactive_roi} plots Dice scores achieved by \texttt{KM-LogUnif} for several ROIs across different values of $\lambda$, densely sampled in $[0, 10]$.
We observe that the optimal Dice for a given ROI occurs at different $\lambda$ values, and furthermore that the profiles of these curves can be quite different.
% We believe that this result highlights one of the main advantages of the proposed method - namely, the ability for the user to choose the appropriate amount of deformation at test time.
The upshot is that since computing registrations with different $\lambda$ values is efficient, an end user can quickly interrogate many optimal registrations, each of which can be optimal for a different ROI.  
% With baseline models, these registrations and their variations would be completely missed and end users would be stuck with one registration. 

\subsubsection{\modelname~can be made robust to noise}
\label{sec:robust_experiments}
\begin{figure}[t]
\centering
\includegraphics[width=\linewidth]{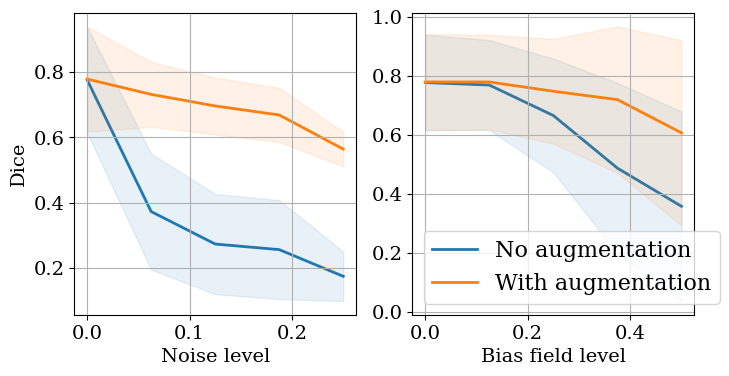}
   \caption{Comparison of KeyMorph performance trained with and without data augmentation (random Gaussian noise and bias field augmentation). KeyMorph achieves adequate robustness by applying augmentations during training. Shaded regions denote standard deviation.}
\label{fig:augmentation}
\end{figure}
We explore whether \modelname~can be made robust to noise factors like bias fields and random Gaussian noise through data augmentation during training.
For random Gaussian noise, we uniformly sample the noise variance in [0, 0.25]. 
For bias field augmentation, we randomly sample the maximum magnitude of polynomial coefficients uniformly in [0, 0.5]. 
We evaluate the models on test subjects over the same range of noise and bias field parameters. 
Fig.~\ref{fig:augmentation} shows that the model can achieve adequate robustness by applying augmentations during training.
In Figs.~\ref{fig:qualitative-augnoise} and~\ref{fig:qualitative-augbias}, we depict qualitative examples corresponding to different degrees of these augmentations.

\subsection{Keypoint Analysis}
\label{sec:experiment3}

\subsubsection{Visualizing keypoints} 
\begin{figure}[t!]
\centering
\includegraphics[width=0.9\linewidth]{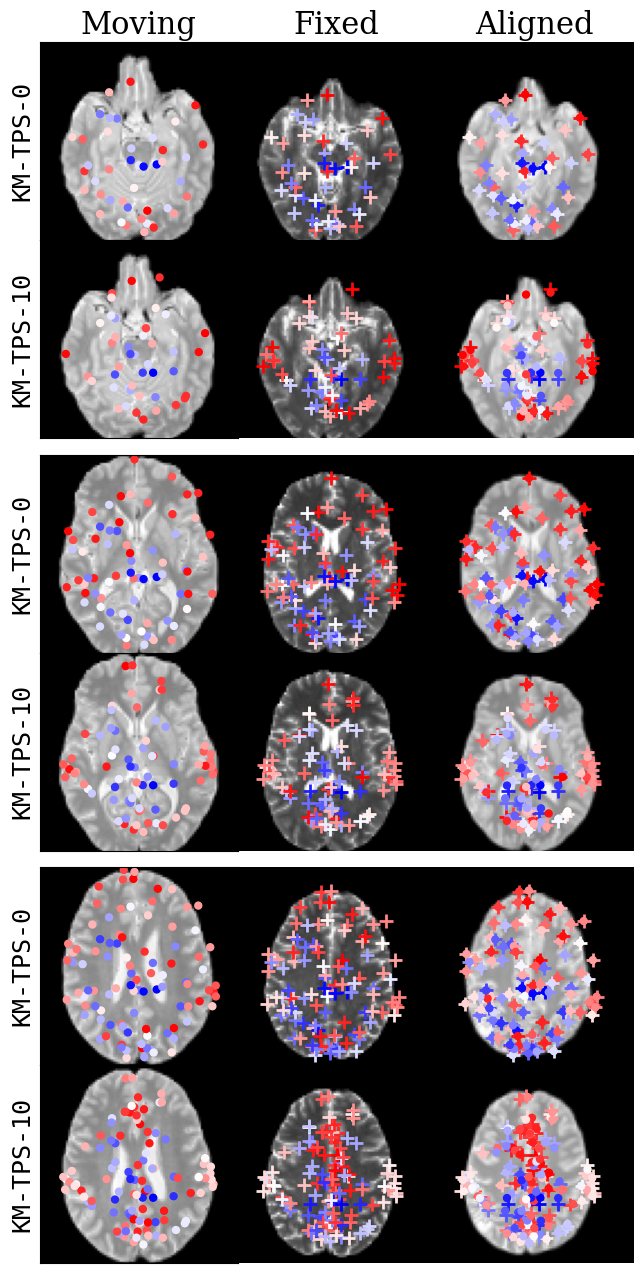}
   \caption{Keypoint visualization for a representative moving and fixed volume pair. For each slice, keypoints that are within 10 voxels of that slice are shown. Colors of keypoints reflect the distance of the keypoint to the visualized slice. For each slice, first row is \texttt{KM-TPS-0}, and second row is \texttt{KM-TPS-10}. Average distance between fixed (crosses) and aligned (dots) keypoints is $0$ voxels for $\lambda=0$ and $16.79$ voxels for $\lambda=10$.}
\label{fig:keypoint_slabs}
\end{figure}
In contrast to existing models that compute the transformation parameters using a “black-box” neural network, we can investigate the keypoints that \modelname~learns to drive the alignment. 
Fig.~\ref{fig:keypoint_slabs} depicts the keypoints for a moving and fixed subject pair across three slices of the volume and for 2 model variants, \texttt{KM-TPS-0} and \texttt{KM-TPS-10}.  
%The color of the keypoints represents depth with respect to the slab.
The ``Aligned" slices show both warped (dots) and fixed (crosses) points.
As expected, we observe that a TPS transformation with $\lambda=0$ exactly aligns the moving and fixed points; this exact interpolation is relaxed for $\lambda=10$.

Keypoint locations are trained end-to-end without explicit annotations.
Therefore, we are interested in the effect of the transformation on the learned keypoint locations.
We observe that the \texttt{KM-TPS-0} keypoints are more evenly distributed over the volume, whereas \texttt{KM-TPS-10} keypoints have a higher degree of regularity and clustering.
We attribute this to the fact that at low values of $\lambda$, each keypoint is a separate degree of freedom and can locally influence the deformation, so spreading out the points maximizes utility.
On the other hand, TPS with $\lambda=10$ is ``affine-like" and has a higher restriction on its local deformation; thus, the model learns to cluster the points in certain regions, which are largely subcortical;
we conjecture that the variability across subjects is low in these regions.

% \subsubsection{Across-Subject Consistency}
% Visually across subjects, we find that the final learned keypoints correspond to the same anatomical region in different subjects and modalities. 
% To quantify this, 

\subsubsection{Keypoints are consistent across subject}
We are interested in whether the model detects similar keypoint locations for volumes of the same subject but different modality, which we refer to as same-subject consistency.
Note that same-subject consistency is explicitly optimized as a loss term in unsupervised variants. 
For supervised variants, consistency is implicit since multi-modal image pairs are presented during training.

\begin{figure}[t]
\centering
\includegraphics[width=0.8\linewidth]{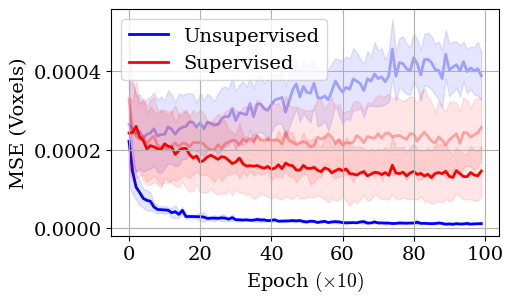}
   \caption{Same-subject keypoint deviation across training, in terms of MSE. Bold lines are standard MSE and Dice models as described in Section.~\ref{sec:variants}. Faint lines are ablated versions of the corresponding model: faint blue is unsupervised training without the same-subject keypoint deviation loss term, and faint red is supervised training but presenting only same-modality pairs in Eq.~\eqref{eq:objective}. Shaded regions denote standard deviation.}
\label{fig:plot_intrascan}
\end{figure}

In Fig.~\ref{fig:plot_intrascan}, we plot the average mean-squared-error (MSE) between the keypoints detected for 3 different combinations of modalities (T1 vs. T2, T1 vs. PD, and T2 vs. PD) across training.
Bold lines denote standard supervised and unsupervised variants, and we observe that both models have decreasing keypoint deviation over training.
Explicit encouragement of consistency in the unsupervised case leads to a higher degree of consistency than implicit encouragement in the supervised case.

In Fig.~\ref{fig:plot_intrascan}, we also investigate ablated versions of both supervised and unsupervised model variants, denoted by the corresponding color in faint lines.
For unsupervised, we remove the explicit keypoint deviation loss.
For supervised, we present only same-modality pairs during training.
We notice that both ablations lead to worse results in terms of same-subject consistency.

\subsubsection{Performance improves with increased keypoints}
\begin{figure}[t]
\centering
\includegraphics[width=0.9\linewidth]{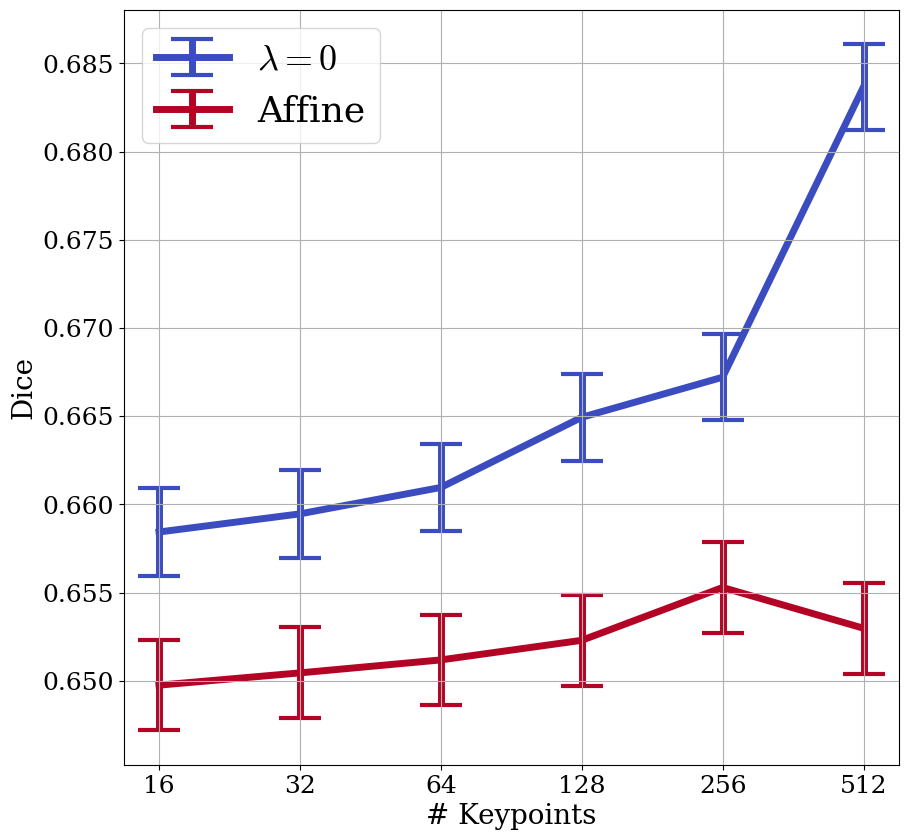}
   \caption{Performance for supervised models over varying number of keypoints. \texttt{KM-TPS-0} (blue) has a higher boost in performance with more keypoints as compared to \texttt{KM-Aff} (red). Error bars denote standard deviation.}
\label{fig:keypoint_box}
\end{figure}
We examine the effect of the number of keypoints used for alignment across different transformations.
We trained supervised \modelname~model variants with 16, 32, 64, 128, 256, and 512 keypoints. 
Fig.~\ref{fig:keypoint_box} illustrates that while increasing the number of keypoints leads to a performance boost in general, the performance boost with more keypoints is higher for \texttt{KM-TPS-0} than for \texttt{KM-Aff} models.
We attribute this to the fact that transformations with higher degrees of nonlinearity benefit more from more degrees of freedom.
For affine transformations, more keypoints is largely redundant since the transformation applies globally.
Indeed, we visually observe that higher numbers of keypoints leads to a higher amount of clustering and overlap of these keypoints for these transformation models (as can be gleaned from Fig.~\ref{fig:keypoint_slabs}).

A current limitation of~\modelname~is the inability to train a higher number of keypoints due to computational constraints. 
We believe more keypoints for nonlinear transformations will lead to even better performance, as evidenced by the blue line in Fig.~\ref{fig:keypoint_box}.
Future research may investigate more memory-efficient keypoint architectures or leverage training strategies which admit a higher number of keypoints to be learned.

\subsubsection{Keypoint extractor is repeatable}
\begin{figure}[t]
\centering
\includegraphics[width=\linewidth]{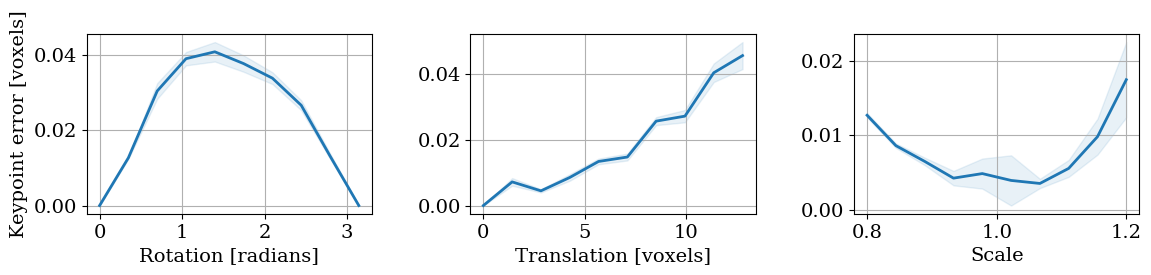}
   \caption{Repeatability of keypoints with respect to transformation. The given transformation is applied to both an image and its corresponding extracted keypoints. Then, keypoints are extracted from the transformed image and compared against the directly transformed keypoints in terms of MSE. In particular, we observe that applying large transformations to the input image (e.g. 180 degree rotations, 15+ voxel translations, or 1.2x scaling) leads to only sub-voxel error in the predicted keypoints. Shaded regions denote standard deviation.}
\label{fig:repeatability}
\end{figure}
In \modelname, robustness to large misalignments is achieved if the keypoint extractor is equivariant with respect to input image deformation.
To verify this, we apply a given transformation to both an image and its corresponding extracted keypoints. 
Then, we extract keypoints from the transformed image and compare the resulting keypoints with the directly transformed keypoints in terms of MSE. 
These results are plotted in Fig.~\ref{fig:repeatability}.
In particular, we observe that applying large transformations to the input image (e.g. 180 degree rotations, 15+ voxel translations, or 1.2x scaling) leads to only sub-voxel error in the predicted keypoints. 
Furthermore, our model readily admits the adoption of any variety of neural architectures which are equivariant under some class of transformations (see, for example,~\citep{zhang2019making}).

\subsubsection{Subjects are discriminable via keypoints}
\begin{figure}[t]
\centering
\includegraphics[width=\linewidth]{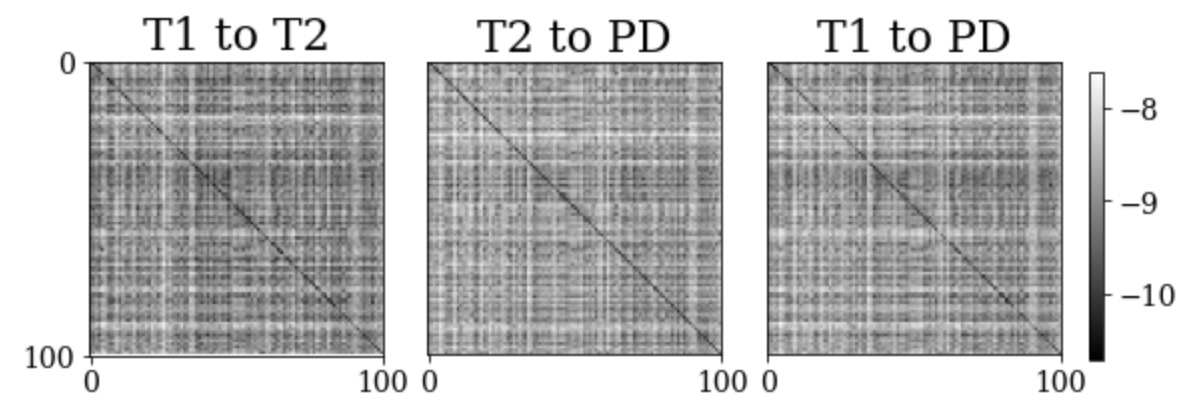}
   \caption{Log of keypoint error matrices across modality. Each row is the MSE of registered keypoints between one subject and every other subject. Matrix takes low value along the diagonal, indicating that keypoints extracted from the same subject are very similar across modalities; that is, a subject is discriminable on the basis of their extracted keypoints.}
\label{fig:discriminability}
\end{figure}
\begin{table}[t]
\centering
\begin{tabular}{lccc}
\hline
$\times 10^{-5}$                     & \textit{T1 to T2} & \textit{T2 to PD} & \textit{T1 to PD} \\ \hline 
\textit{Same subject}           & 7.00$\pm$3.29   & 4.87$\pm$1.85    & 6.46$\pm$3.75                           \\
\textit{Different subject}      & 16.48$\pm$6.35   & 15.99$\pm$6.36   & 15.78$\pm$6.43                         \\
\hline
\end{tabular}%
\caption{Subject-wise discriminability of predicted keypoints. Mean $\pm$ standard deviation of Euclidean distance of keypoints between different modalities, split between same subject and different subject. Same subject keypoint distance is much lower, indicating subjects are discriminable on the basis of keypoints.}
\label{tab:discriminability_table}
\end{table}
If extracted keypoints are anatomically consistent, we would expect a subject to be discriminable on the basis of their keypoints (i.e. that their keypoints are ``personalized").
Put another way, a subject's keypoints extracted from modality 1 should be closest to its keypoints extracted from modality 2, compared to all other subjects' keypoints extracted from modality 2 (and assuming all images are registered to a standardized space).

Fig.~\ref{fig:discriminability} shows error matrices between keypoints across all 100 subjects in the test set, for all 3 modality pairs (T1 to T2, T1 to PD, and T2 to PD).
To generate these error matrices, we first extract keypoints from all test subjects for all 3 modalities (T1, T2, and PD). 
For each subject and a given modality, we perform affine registration of every other subject (including itself) in a different modality to that first subject. 
Then, we compute the MSE of registered keypoints between that subject and every other subject, which constitutes a single row in the matrix. 
We observe that this matrix takes low value along the diagonal, indicating that keypoints extracted from the same subject are very similar across modalities; that is, a subject is discriminable on the basis of their extracted keypoints.
In Table~\ref{tab:discriminability_table}, we quantify this by averaging the errors over same subject and different subject (essentially, averaging over diagonal and non-diagonal entries).

\section{Conclusion}
We presented~\modelname, a deep learning-based image registration method that uses corresponding keypoints to derive the optimal transformation that align the images.
By unifying image registration and keypoint detection, we can train a model that finds matching keypoints useful for aligning images. 
In addition, our method has the capability of having a higher degree of robustness to large misalignments, interpretablity, controllability, and invariances inherent to the registration problem.
We empirically demonstrate these capabilities on a large, real-world dataset of multi-modal brain MRI volumes.

\section*{Acknowledgments}
Funding for this project was in part provided by the NIH grants R01AG053949, R01AG064027 and R01AG070988, and the NSF CAREER 1748377 grant.

%%Harvard
\bibliographystyle{model2-names.bst}\biboptions{authoryear}
\bibliography{refs}

\clearpage
\appendix
\onecolumn
\section{Derivation of Eq.~\eqref{eq:closeform}}
\label{appendix:proof}
We wish to find the optimal affine transformation~$\bm{A} \in \mathbb{R}^{D\times (D+1)}$ which minimizes:
\begin{align*} 
\mathcal{L} &= \sum_{i=1}^N \left(\bm{A} \tilde{\bm{p}}^{(i)} - \bm{q}^{(i)}\right)^2 \\
            &=\left \| \bm{A} \tilde{\boldsymbol{P}}- \boldsymbol{Q}\right \|_{F},
\end{align*}
where~$\norm{\cdot}_F$ denotes the Frobenius norm.
Taking the derivative with respect to~$\bm{A}$ and setting the result to zero, we obtain:
\begin{align*}
    \frac{\partial \mathcal{L}}{\partial \bm{A}} &=  (\bm{A}\tilde{\boldsymbol{P}} - \boldsymbol{Q}) \tilde{\boldsymbol{P}}^{T} = \boldsymbol{0} \\
    &\implies  \bm{A}\tilde{\boldsymbol{P}}\tilde{\boldsymbol{P}}^{T}   = \boldsymbol{Q}\tilde{\boldsymbol{P}}^{T} \\
    &\implies  \bm{A} = \boldsymbol{Q}\tilde{\boldsymbol{P}}^{T} (\tilde{\boldsymbol{P}}\tilde{\boldsymbol{P}}^{T})^{-1}.
\end{align*}
\section{Additional Figures}
\noindent%
\begin{minipage}{\textwidth}% to keep image and caption on one page
\makebox[\textwidth]{%        to center the image
  \includegraphics[keepaspectratio=true,scale=0.5]{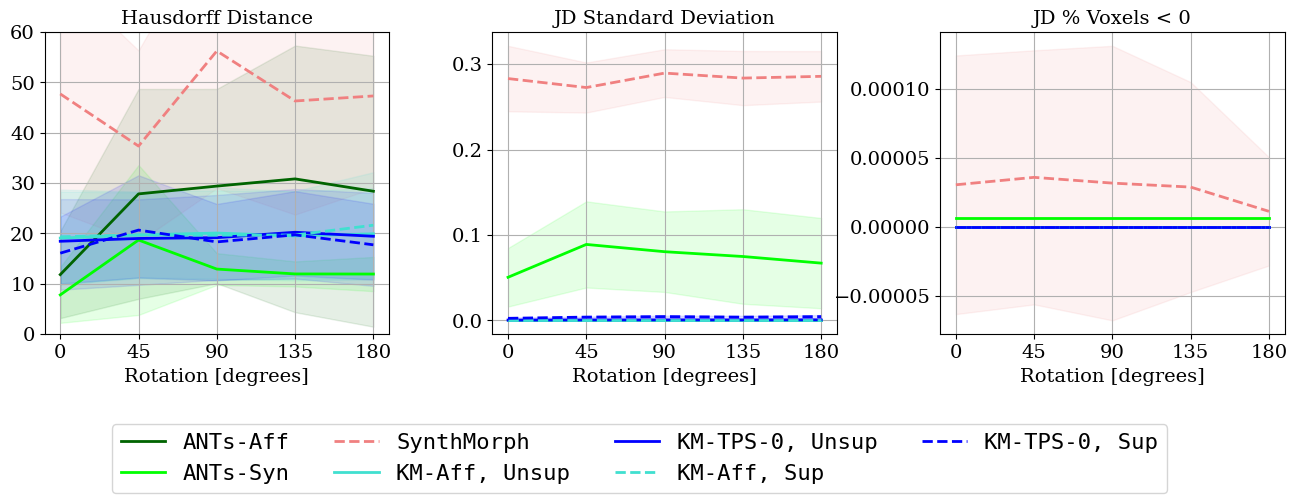}}
\captionof{figure}{KeyMorph and baseline model performance across rotation angle for various metrics. JD is short for Jacobian determinant. In the last panel, KeyMorph variants are $0$ across all rotation angles. Hausdorff Distance is computed with respect to the brain surface only. Shaded regions denote standard deviation.}\label{fig:hd_J_metrics}%      only if needed  
\end{minipage}
\begin{figure*}[t]
\centering
\includegraphics[width=0.65\linewidth]{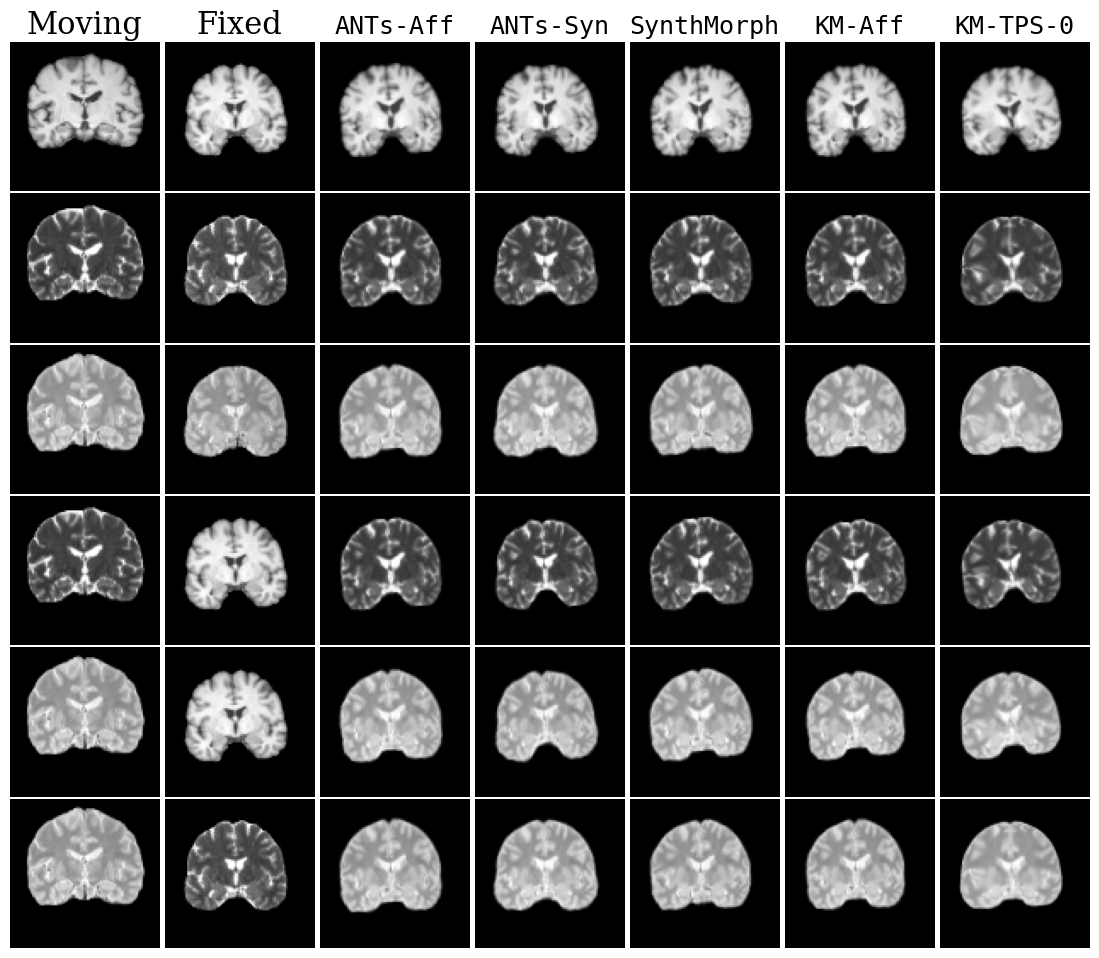}
   \caption{Qualitative results for all models. Moving image is not misaligned. DLIR results are omitted to save space.}
\label{fig:qualitative-rot0}
\end{figure*}
\begin{figure*}[t]
\centering
\includegraphics[width=0.65\linewidth]{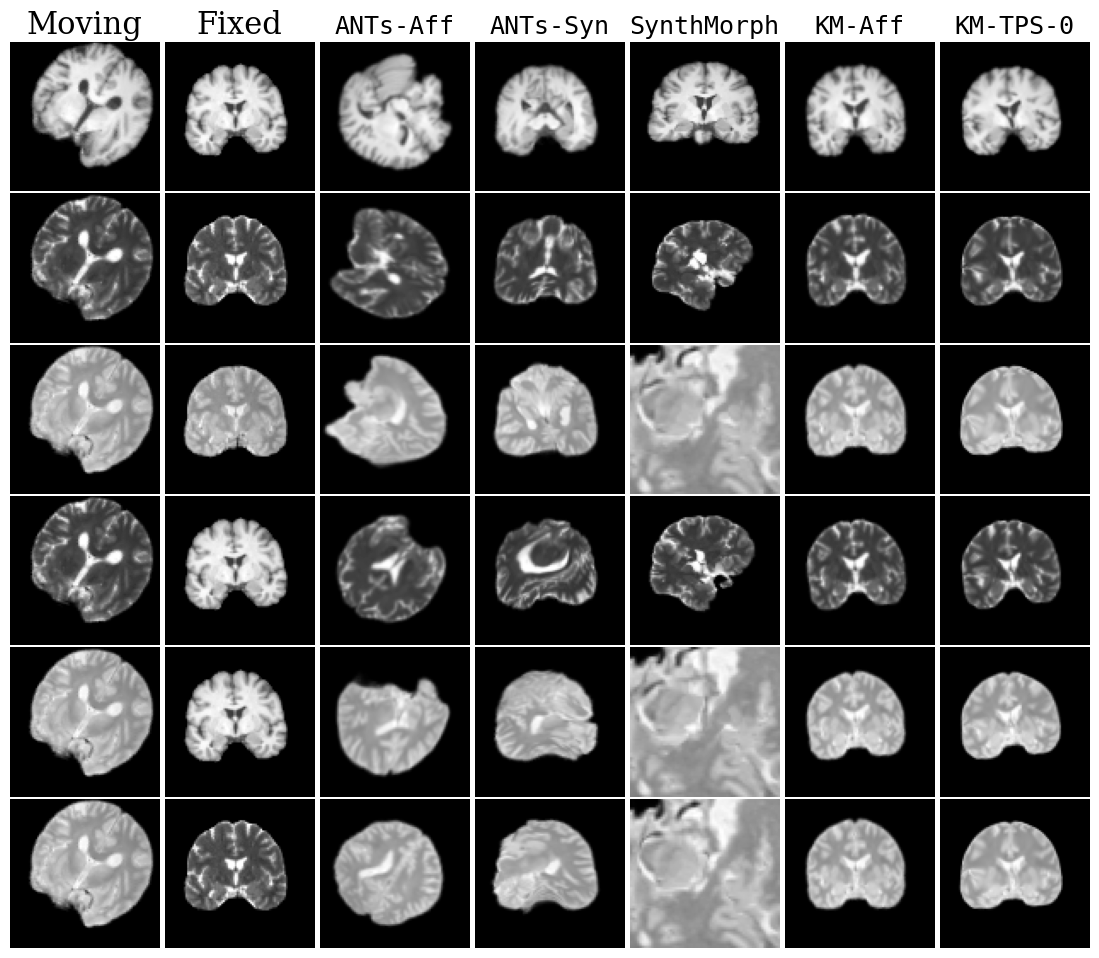}
   \caption{Qualitative results for all models. Moving image is misaligned by 45\degree~in each dimension. DLIR results are omitted to save space.}
\label{fig:qualitative-rot90}
\end{figure*}

\begin{figure*}[t]
\centering
\includegraphics[width=0.65\linewidth]{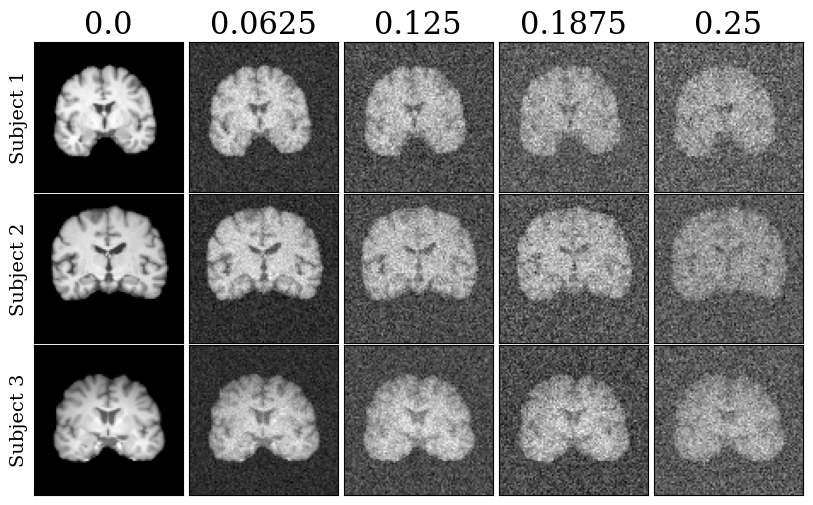}
   \caption{Qualitative examples for additive Gaussian noise experiments in Section~\ref{sec:robust_experiments}. Rows depict different subjects. Columns correspond to varying noise standard deviation.}
\label{fig:qualitative-augnoise}
\end{figure*}

\begin{figure*}[t]
\centering
\includegraphics[width=0.65\linewidth]{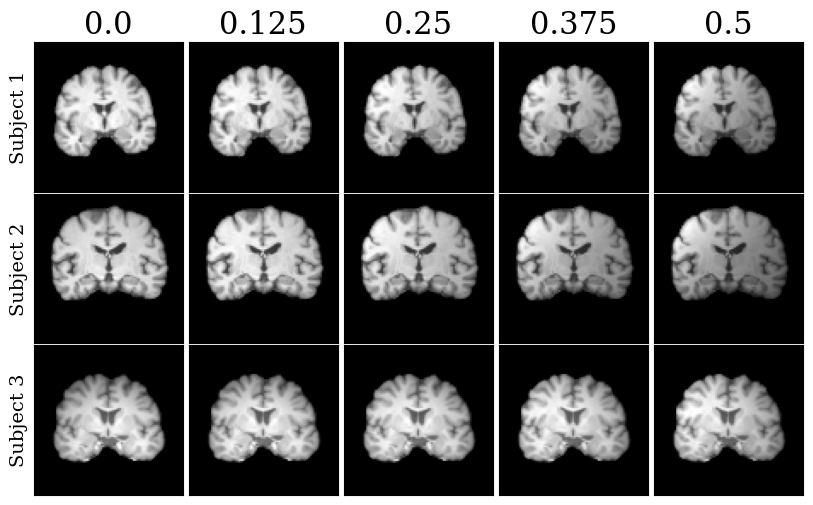}
   \caption{Qualitative examples for bias field augmentation experiments in Section~\ref{sec:robust_experiments}. Rows depict different subjects. Columns correspond to varying magnitude of polynomial coefficients.}
\label{fig:qualitative-augbias}
\end{figure*}

\clearpage
\begin{table*}
\centering
\scalebox{0.72}{
\begin{tabular}{l|l|ccccccccc}
\hline
\multirow{2}{*}{T}           & \multirow{2}{*}{Model}         & \multicolumn{9}{c}{Dice Score}                                                                                                                                                                                                                                                             \\ \cline{3-11} 
                             &                                & T1$\rightarrow$T1        & T2$\rightarrow$T1        & \multicolumn{1}{c|}{PD$\rightarrow$T1}        & T1$\rightarrow$T2        & T2$\rightarrow$T2        & \multicolumn{1}{c|}{PD$\rightarrow$T2}        & T1$\rightarrow$PD        & T2$\rightarrow$PD        & PD$\rightarrow$PD        \\ \hline
\multirow{9}{*}{\rotatebox[origin=c]{90}{rotation}}    &~No registration     & 20.28$\pm$19.80           & 12.67$\pm$12.52                        & \multicolumn{1}{c|}{18.68$\pm$15.56}                        & 12.67$\pm$12.52                        & 24.02$\pm$22.28          & \multicolumn{1}{c|}{18.46$\pm$16.99}                        & 18.68$\pm$15.56                        & 18.46$\pm$16.99                        & 19.67$\pm$19.65          \\
                             &~$\texttt{DLIR-Uni-180-unsup}$     & 26.8$\pm$14.84           & -                        & \multicolumn{1}{c|}{-}                        & -                        & 26.86$\pm$15.01          & \multicolumn{1}{c|}{-}                        & -                        & -                        & 29.36$\pm$17.01          \\
                             &~$\texttt{DLIR-Uni-35-unsup}$      & 51.33$\pm$7.95           & -                        & \multicolumn{1}{c|}{-}                        & -                        & 53.88$\pm$7.2            & \multicolumn{1}{c|}{-}                        & -                        & -                        & 53.57$\pm$6.35           \\
                             &~$\texttt{DLIR-Uni-180-sup}$    & 38.64$\pm$14.65          & -                        & \multicolumn{1}{c|}{-}                        & -                        & 28.59$\pm$12.62          & \multicolumn{1}{c|}{-}                        & -                        & -                        & 35.24$\pm$13.98          \\
                             &~$\texttt{DLIR-Uni-35-sup}$     & 52.57$\pm$7.81           & -                        & \multicolumn{1}{c|}{-}                        & -                        & 53.41$\pm$9.33           & \multicolumn{1}{c|}{-}                        & -                        & -                        & 54.14$\pm$8.03           \\
                             &~$\texttt{DLIR-Multi-180-sup}$  & 21.68$\pm$11.06          & 21.78$\pm$11.42          & \multicolumn{1}{c|}{22.0$\pm$11.46}           & 21.32$\pm$10.6           & 21.88$\pm$11.48          & \multicolumn{1}{c|}{21.99$\pm$11.37}          & 21.22$\pm$10.67          & 21.78$\pm$11.5           & 22.09$\pm$11.62          \\
                             &~$\texttt{DLIR-Multi-35-sup}$   & 49.84$\pm$8.48           & 48.98$\pm$8.16           & \multicolumn{1}{c|}{49.37$\pm$8.41}           & 49.27$\pm$8.33           & 49.42$\pm$8.21           & \multicolumn{1}{c|}{49.41$\pm$8.36}           & 49.52$\pm$8.31             & 49.15$\pm$8.11           & 50.28$\pm$8.62            \\
                             &~$\texttt{ANTs-Aff}$                & 48.07$\pm$26.14          & 44.32$\pm$25.34          & \multicolumn{1}{c|}{42.27$\pm$25.0}           & 43.48$\pm$25.23          & 45.06$\pm$25.59          & \multicolumn{1}{c|}{42.69$\pm$25.04}          & 40.56$\pm$25.01          & 43.51$\pm$24.89          & 44.66$\pm$25.84          \\
                             &~$\texttt{ANTs-Syn}$                & 49.33$\pm$24.39          & 46.02$\pm$22.82          & \multicolumn{1}{c|}{43.82$\pm$23.18}           & 45.13$\pm$22.07          & 46.82$\pm$26.35          & \multicolumn{1}{c|}{43.23$\pm$23.63}          & 42.69$\pm$25.60          & 45.83$\pm$22.90          & 46.03$\pm$23.29          \\
                             &~$\texttt{SynthMorph}$  & 67.20$\pm$22.42           & 47.79$\pm$18.55           & \multicolumn{1}{c|}{52.18$\pm$21.98}            & 48.48$\pm$18.68           & 46.64$\pm$17.39           & \multicolumn{1}{c|}{44.15$\pm$18.53}           & 51.98$\pm$22.43           & 45.62$\pm$18.69           & 45.42$\pm$20.91           \\
                             &~$\texttt{KM-Aff-unsup}$  & 64.35$\pm$11.18           & 57.68$\pm$10.78           & \multicolumn{1}{c|}{60.08$\pm$9.51}            & 57.85$\pm$10.26           & 61.18$\pm$15.77           & \multicolumn{1}{c|}{58.68$\pm$12.01}           & 59.05$\pm$8.42           & 59.04$\pm$11.45           & 61.19$\pm$12.97           \\
                             &~$\texttt{KM-Aff-sup}$ &~67.56$\pm$9.39 &~61.69$\pm$5.34 & \multicolumn{1}{c|}{62.46$\pm$6.42}  &~61.40$\pm$10.62 &~63.16$\pm$5.3  & \multicolumn{1}{c|}{61.11$\pm$4.83} &~63.44$\pm$4.87 &~61.11$\pm$5.11 &~63.32$\pm$5.26 \\ 
                             &~$\texttt{KM-TPS-0-unsup}$  & 61.33$\pm$6.56           & 53.23$\pm$5.99           & \multicolumn{1}{c|}{53.65$\pm$5.78}            & 52.60$\pm$6.07           & 59.62$\pm$6.28           & \multicolumn{1}{c|}{55.21$\pm$6.57}           & 54.22$\pm$6.01           & 54.77$\pm$5.68           & 56.72$\pm$6.21           \\
                             &~$\texttt{KM-TPS-0-sup}$ &~$\mathbf{70.54\pm5.11}$ &~$\mathbf{65.29\pm5.04}$ & \multicolumn{1}{c|}{$\mathbf{66.24\pm4.82}$}  &~$\mathbf{65.29\pm4.96}$ &~$\mathbf{65.45\pm5.36}$  & \multicolumn{1}{c|}{$\mathbf{63.89\pm4.83}$} &~$\mathbf{66.24\pm4.87}$ &~$\mathbf{63.56\pm5.11}$ &~$\mathbf{66.24\pm5.26}$ \\ \hline
\multirow{9}{*}{\rotatebox[origin=c]{90}{scaling}}    &~No registration     & 22.33$\pm$14.24           & 12.28$\pm$13.28                        & \multicolumn{1}{c|}{14.24$\pm$15.94}                        & 12.28$\pm$13.28                        & 25.23$\pm$19.02          & \multicolumn{1}{c|}{19.24$\pm$12.34}                        & 14.24$\pm$15.94                        & 19.24$\pm$12.34                        & 22.34$\pm$14.24          \\
                             &~$\texttt{DLIR-Uni-180-unsup}$     & 45.97$\pm$8.07           & -                        & \multicolumn{1}{c|}{-}                        & -                        & 45.96$\pm$8.46           & \multicolumn{1}{c|}{-}                        & -                        & -                        & 53.47$\pm$7.32           \\
                             &~$\texttt{DLIR-Uni-35-unsup}$      & 53.32$\pm$7.26           & -                        & \multicolumn{1}{c|}{-}                        & -                        & 54.46$\pm$7.37           & \multicolumn{1}{c|}{-}                        & -                        & -                        & 53.37$\pm$5.98           \\
                             &~$\texttt{DLIR-Uni-180-sup}$    & 44.23$\pm$9.13           & -                        & \multicolumn{1}{c|}{-}                        & -                        & 39.09$\pm$7.48           & \multicolumn{1}{c|}{-}                        & -                        & -                        & 42.91$\pm$7.74           \\
                             &~$\texttt{DLIR-Uni-35-sup}$     & 55.19$\pm$6.95           & -                        & \multicolumn{1}{c|}{-}                        & -                        & 54.92$\pm$7.96           & \multicolumn{1}{c|}{-}                        & -                        & -                        & 57.57$\pm$6.8            \\
                             &~$\texttt{DLIR-Multi-180-sup}$  & 39.12$\pm$9.49           & 41.08$\pm$7.93           & \multicolumn{1}{c|}{41.32$\pm$7.81}           & 38.12$\pm$9.45           & 41.74$\pm$8.43           & \multicolumn{1}{c|}{41.72$\pm$8.18}             & 37.96$\pm$9.93           & 41.48$\pm$8.85           & 42.02$\pm$8.81           \\
                             &~$\texttt{DLIR-Multi-35-sup}$   & 53.43$\pm$6.35           & 52.68$\pm$6.33           & \multicolumn{1}{c|}{52.86$\pm$6.54}           & 52.57$\pm$6.56           & 52.84$\pm$6.98           & \multicolumn{1}{c|}{52.49$\pm$6.71}           & 52.69$\pm$6.37           & 52.38$\pm$6.76           & 53.32$\pm$6.74           \\
                             &~$\texttt{ANTs-Aff}$                & 66.29$\pm$6.19           &~63.42$\pm$5.75   & \multicolumn{1}{c|}{62.51$\pm$5.93}           & 62.7$\pm$5.81            & 63.84$\pm$6.33           & \multicolumn{1}{c|}{62.42$\pm$5.74}           & 61.49$\pm$5.94           & 62.28$\pm$5.7            & 64.29$\pm$6.48           \\
                             &~$\texttt{ANTs-Syn}$                & 66.60$\pm$5.69           &~63.91$\pm$6.10   & \multicolumn{1}{c|}{62.83$\pm$6.48}           & 63.0$\pm$5.30            & 63.98$\pm$5.78           & \multicolumn{1}{c|}{62.70$\pm$4.61}           & 61.26$\pm$6.27           & 63.06$\pm$5.37            & 64.62$\pm$5.84           \\
                             &~$\texttt{SynthMorph}$                & 68.18$\pm$7.37           &~$\mathbf{64.80\pm6.21}$   & \multicolumn{1}{c|}{63.84$\pm$6.13}           & 64.21$\pm$5.28            & 65.39$\pm$7.26           & \multicolumn{1}{c|}{64.70$\pm$6.30}           & 63.18$\pm$6.52           & 63.27$\pm$5.14            & 65.13$\pm$6.82           \\
                             &~$\texttt{KM-Aff-unsup}$  & 63.0$\pm$6.79            & 59.23$\pm$6.02           & \multicolumn{1}{c|}{59.7$\pm$6.65}            & 59.35$\pm$6.13           & 61.73$\pm$6.49           & \multicolumn{1}{c|}{60.27$\pm$6.01}           & 59.66$\pm$6.28           & 60.16$\pm$5.62           & 62.8$\pm$6.64            \\
                             &~$\texttt{KM-Aff-sup}$ &~66.51$\pm$5.24 & 63.25$\pm$4.86           & \multicolumn{1}{c|}{63.35$\pm$4.6}  &~63.33$\pm$4.98 &~64.16$\pm$5.35 & \multicolumn{1}{c|}{62.7$\pm$4.55}  &63.37$\pm$5.0  &~62.77$\pm$4.81 &~64.85$\pm$5.42 \\ 
                             &~$\texttt{KM-TPS-0-unsup}$  & 62.53$\pm$7.30            & 57.51$\pm$6.83           & \multicolumn{1}{c|}{57.38$\pm$6.13}            & 56.87$\pm$7.39           & 58.79$\pm$6.98           & \multicolumn{1}{c|}{58.18$\pm$6.94}           & 56.87$\pm$6.10           & 57.27$\pm$7.32           & 60.92$\pm$5.97            \\
                             &~$\texttt{KM-TPS-0-sup}$ &~$\mathbf{69.23\pm 6.31}$ & 63.83$\pm$4.90           & \multicolumn{1}{c|}{$\mathbf{65.72\pm 5.17}$}  &~$\mathbf{66.29\pm 5.42}$ &~$\mathbf{66.75\pm 4.98}$ & \multicolumn{1}{c|}{$\mathbf{64.84\pm 5.07}$}  &~$\mathbf{65.28\pm 5.82}$  &~$\mathbf{65.81\pm 4.96}$ &~$\mathbf{66.12\pm 5.03}$ \\ \hline 
\multirow{9}{*}{\rotatebox[origin=c]{90}{translation}}    &~No registration     & 14.38$\pm$16.24           & 12.64$\pm$15.90                        & \multicolumn{1}{c|}{15.30$\pm$15.22}                        & 12.64$\pm$15.90                        & 19.23$\pm$20.30          & \multicolumn{1}{c|}{17.36$\pm$12.41}                        & 15.30$\pm$15.22                        & 17.36$\pm$12.41                        & 11.34$\pm$19.10          \\
                             &~$\texttt{DLIR-Uni-180-unsup}$     & 45.88$\pm$8.12           & -                        & \multicolumn{1}{c|}{-}                        & -                        & 45.57$\pm$8.25           & \multicolumn{1}{c|}{-}                        & -                        & -                        & 53.14$\pm$7.73           \\
                             &~$\texttt{DLIR-Uni-35-unsup}$      & 52.8$\pm$7.84            & -                        & \multicolumn{1}{c|}{-}                        & -                        & 54.43$\pm$7.56           & \multicolumn{1}{c|}{-}                        & -                        & -                        & 53.32$\pm$6.33           \\
                             &~$\texttt{DLIR-Uni-180-sup}$    & 45.05$\pm$8.84           & -                        & \multicolumn{1}{c|}{-}                        & -                        & 38.59$\pm$7.28           & \multicolumn{1}{c|}{-}                        & -                        & -                        & 43.09$\pm$8.07           \\
                             &~$\texttt{DLIR-Uni-35-sup}$     & 55.26$\pm$6.97           & -                        & \multicolumn{1}{c|}{-}                        & -                        & 54.77$\pm$7.97           & \multicolumn{1}{c|}{-}                        & -                        & -                        & 57.65$\pm$6.69           \\
                             &~$\texttt{DLIR-Multi-180-sup}$  & 37.52$\pm$9.69           & 38.78$\pm$8.52           & \multicolumn{1}{c|}{39.25$\pm$8.35}           & 36.32$\pm$9.57           & 39.1$\pm$9.18            & \multicolumn{1}{c|}{39.3$\pm$8.79}            & 36.18$\pm$10.03          & 38.92$\pm$9.5            & 39.7$\pm$9.38            \\
                             &~$\texttt{DLIR-Multi-35-sup}$   & 53.41$\pm$6.17           & 52.38$\pm$6.07           & \multicolumn{1}{c|}{52.67$\pm$6.5}            & 52.84$\pm$6.27           & 52.78$\pm$6.58           & \multicolumn{1}{c|}{52.6$\pm$6.68}            & 52.88$\pm$6.49           & 52.33$\pm$6.62           & 53.33$\pm$6.93           \\
                             &~$\texttt{ANTs-Aff}$                & 66.34$\pm$6.37           & 63.45$\pm$5.7            & \multicolumn{1}{c|}{62.49$\pm$5.9}            & 62.79$\pm$5.81           & 63.94$\pm$6.46           & \multicolumn{1}{c|}{62.44$\pm$5.73}           & 61.5$\pm$5.97            & 62.34$\pm$5.72           & 64.3$\pm$6.62            \\
                             &~$\texttt{ANTs-Syn}$                & 66.70$\pm$5.83           & 63.81$\pm$5.35           & \multicolumn{1}{c|}{62.18$\pm$5.30}            & 63.01$\pm$6.04           & 63.53$\pm$5.91           & \multicolumn{1}{c|}{63.30$\pm$5.02}           & 62.06$\pm$4.98            & 62.70$\pm$6.41           & 64.09$\pm$6.06            \\
                             &~$\texttt{SynthMorph}$                & 69.13$\pm$7.71           & 63.45$\pm$5.7            & \multicolumn{1}{c|}{64.20$\pm$6.45}            & 64.31$\pm$6.50           & 65.80$\pm$7.13           & \multicolumn{1}{c|}{64.24$\pm$6.31}           & 63.58$\pm$6.73            & 64.80$\pm$6.30           & 66.58$\pm$7.31            \\
                             &~$\texttt{KM-Aff-unsup}$  & 63.5$\pm$7.09            & 59.5$\pm$5.92            & \multicolumn{1}{c|}{59.84$\pm$6.64}           & 59.86$\pm$6.02           & 62.33$\pm$6.63           & \multicolumn{1}{c|}{60.58$\pm$5.9}            & 60.19$\pm$6.19           & 60.64$\pm$5.5            & 63.28$\pm$6.78           \\
                             &~$\texttt{KM-Aff-sup}$ &~66.99$\pm$5.41 &~63.55$\pm$4.82 & \multicolumn{1}{c|}{63.75$\pm$4.64} &~63.94$\pm$4.71 &~64.79$\pm$5.62 & \multicolumn{1}{c|}{63.34$\pm$4.55} &~63.83$\pm$4.74 &~63.16$\pm$4.86 &~65.42$\pm$5.66 \\ 
                             &~$\texttt{KM-TPS-0-unsup}$  & 62.05$\pm$6.34            & 57.40$\pm$5.11            & \multicolumn{1}{c|}{58.79$\pm$6.14}           & 58.74$\pm$7.31           & 61.30$\pm$7.23           & \multicolumn{1}{c|}{59.94$\pm$5.20}            & 59.22$\pm$5.67           & 59.51$\pm$6.90            & 61.13$\pm$6.18           \\
                             &~$\texttt{KM-TPS-0-sup}$ &~$\mathbf{70.60\pm 4.99}$ &~$\mathbf{66.23\pm 5.85}$ & \multicolumn{1}{c|}{$\mathbf{65.82\pm 5.06}$} &~$\mathbf{66.25\pm 5.13}$ &~$\mathbf{66.82\pm 5.16}$ & \multicolumn{1}{c|}{$\mathbf{66.32\pm 5.10}$} &~$\mathbf{65.30\pm 5.83}$ &~$\mathbf{66.48\pm 5.31}$ &~$\mathbf{68.13\pm 4.98}$ \\ \hline
\end{tabular}%
}
\caption{\label{tab:big-table} Mean performance of all methods $\pm$ standard deviation. The average Dice score is computed across test subject pairs, brain regions, and modalities. The notation~$A \rightarrow B$ refers to registering moving volumes of modality~$A$ to fixed volumes of modality~$B$. Bold numbers highlight the highest Dice score of a task given a transformation shown in the first column T. }
\end{table*}

%

% \section*{Supplementary Material}

% Supplementary material that may be helpful in the review process should
% be prepared and provided as a separate electronic file. That file can
% then be transformed into PDF format and submitted along with the
% manuscript and graphic files to the appropriate editorial office.

\end{document}